\pdfoutput=1

\documentclass[11pt]{article}

\usepackage[final]{coling}

\usepackage{times}
\usepackage{latexsym}

\newcommand{\grammaMT}{\textsc{GrammaMT}}

\usepackage[T1]{fontenc}

\usepackage[utf8]{inputenc}

\usepackage{microtype}

\usepackage{inconsolata}

\usepackage[utf8]{inputenc}

\usepackage{microtype}
\usepackage{graphicx}
\usepackage{amsmath}
\usepackage{amssymb}
\usepackage{booktabs}
\usepackage{comment}
\usepackage{multirow}
\usepackage{cleveref}
\usepackage{algorithm}
\usepackage{algpseudocode}
\usepackage{xspace}
\usepackage[skip=4pt]{caption}
\usepackage{subcaption}
\usepackage[inline]{enumitem}

\usepackage{placeins}

\usepackage{booktabs}
\usepackage{xcolor,colortbl}

\definecolor{Gray}{gray}{0.85}
\newcolumntype{a}{>{\columncolor{Gray}}c}

\newcommand{\everlyn}[1]{\textcolor{magenta}{Eva: #1}}

\title{ \grammaMT \raisebox{-4pt}{\includegraphics[width=0.24in]{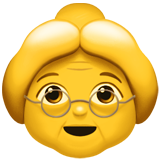}}\hspace{0.9pt}: Improving Machine Translation with Grammar-Informed In-Context Learning }

\author{Rita Ramos$^{\diamond}$\ \ Everlyn Asiko Chimoto$^{\dagger}$\thanks{Work done while at Apple.}  \ \ Maartje ter Hoeve $^{\diamond}$
 \ \ Natalie Schluter $^{\diamond}$
 \\
$^{\diamond}$Apple \\
$^{\dagger}$University of Cape Town, South Africa \\
\texttt{rita\_ramos@apple.com}}

\begin{document}

\maketitle
\begin{abstract}

We introduce \grammaMT, a grammatically-aware prompting approach for machine translation that uses Interlinear Glossed Text (IGT), a common form of linguistic description providing morphological and lexical annotations for source sentences. \grammaMT{} proposes three prompting strategies: gloss-shot, chain-gloss and model-gloss. All are training-free, requiring only a few examples that involve minimal effort to collect, and making them well-suited for low-resource setups. 
Experiments show that \grammaMT{} enhances translation performance on open-source instruction-tuned LLMs for various low- to high-resource languages across three benchmarks: (1) the largest IGT corpus, (2) the challenging 2023 SIGMORPHON Shared Task data over endangered languages, and (3) even in an out-of-domain setting with FLORES.
Moreover, ablation studies reveal that leveraging gloss resources could substantially boost MT performance (by over 17 BLEU points) if LLMs accurately generate or access input sentence glosses.
 

\end{abstract}

\section{Introduction}
\begin{figure*}[!t]
  \centering
    \includegraphics[width=\linewidth]{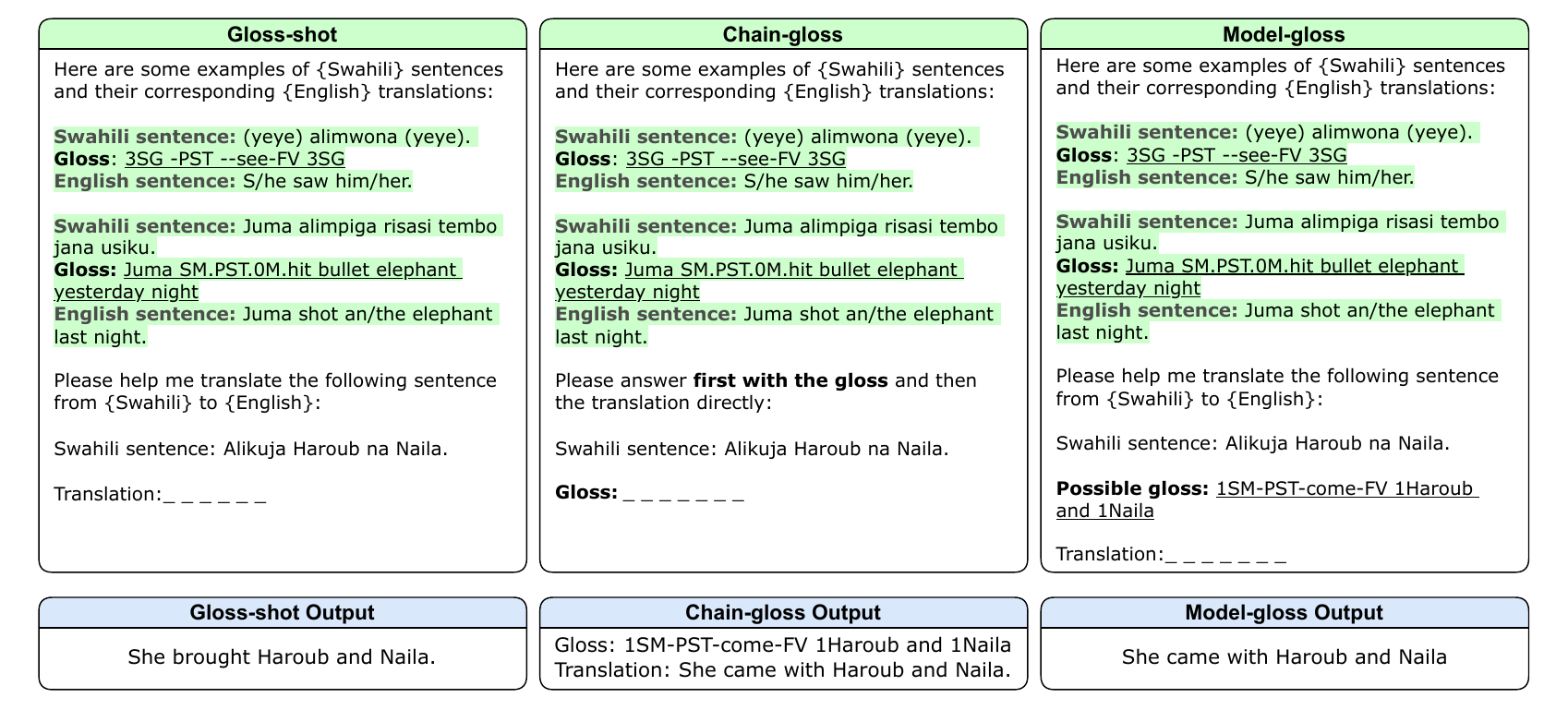}
  \caption{\grammaMT{} augments few-shot learning with Interlinear Gloss Text. In gloss-shot, the LLM is conditioned on translation pairs with source glosses. In chain-gloss, the LLM first generates the gloss before translating. Lastly, in model-gloss, the LLM receives an input gloss from an external gloss generation model.}
  \label{fig:approach}
\vspace{-0.35cm}
\end{figure*}
Large Language Models (LLMs) have taken over the NLP leaderboards~\cite[e.g.,][]{zellers2019hellaswag, hendrycks2020measuring, alpaca_eval}. Training LLMs requires access to a plethora of datasets, a luxury accessible to only a few of the world's most high-resource languages. Consequently, only a sliver of the world's languages have sufficient data for LLMs to achieve these impressive performance gains~\cite{achiam2023gpt, ustun2024aya}.
To leverage the capabilities of these existing, high-resource LLMs in a low-resource context, one needs to design an approach that requires: \begin{enumerate*}[label=(\roman*)]
    \item \textbf{little to no training} (to avoid overfitting and catastrophic forgetting), 
    \item only a \textbf{small amount of data}, and/or 
    \item \textbf{ease in data collection}.
\end{enumerate*}

Recent studies have shown the capability of LLMs to perform complex tasks, when provided with only a small amount of high quality language data. This data comes in the form of instruction-answer pairs for instruction fine-tuning~\cite[e.g,][]{li2023self, yuan2024self} or in the form of high quality prompts~\cite[e.g,][]{wei2022chain}. For example, for machine translation of languages unseen during training, performance gains have been achieved by only providing a dictionary and grammar book for the unseen languages as input to the LLM~\citep{tanzer2024a,zhang2024hire}.


Motivated by these results and the three requirements above, we propose \grammaMT, an in-context learning approach that leverages grammatical information from Interlinear Glossed Text (IGT) to improve machine translation in both low and high-resource settings. IGT is a triplet of source sentence, gloss, and target translation, commonly used by grammarians and linguists in linguistic description. The gloss represents the source sentence as a sequence of morphological and lexical annotations, as illustrated in Figure \ref{fig:approach}.



~\grammaMT{} introduces three prompting strategies that augment  few-shot machine translation using annotated glosses: \begin{enumerate*}[label=(\roman*)]
\item \textbf{gloss-shot}, 
\item \textbf{chain-gloss} and 
\item \textbf{model-gloss}
\end{enumerate*}. In gloss-shot, the LLM is prompted with examples pairing source sentences both with their translations and their glosses. In chain-gloss, the LLM first generates a gloss of the source sentence 
before translating. Model-gloss uses an external gloss model to generate the gloss, reducing the risk of incorrect glosses in chain-gloss, especially if a specialised gloss model is available for the target language. Importantly, \grammaMT{} adheres to all three of the above design requirements as follows.


%
\paragraph{Training-free.} \grammaMT{} works by simply prompting an LLM with a grammatical demonstration. This is especially important in low-resource settings, where sufficiently large training datasets are scarce, but minimal linguistic annotations exist or can be obtained. By incorporating linguistic knowledge directly into the prompt, we effectively leverage limited linguistic data that would otherwise be insufficient for fine-tuning an LLM.
\paragraph{Small number of examples.} 

\grammaMT{} needs only a small number of grammatical annotations (e.g., 21 interlinear glosses examples). This differs from other few-shot methods, which depend on acquiring large data stores to gather relevant samples (e.g. retrieval-augmentation) or extensive resources like dictionaries or grammar chapters.
\paragraph{Ease of collection.} Unlike chain-of-thought examples \cite{wei_etal2022}, which require costly and subjective human engineering to break down machine translation into smaller steps, \grammaMT{} relies on basic gloss notation.  These annotations are more straightforward--easier to either manually collect
in low-resource settings, or can be sourced from grammar books or automatically generated \cite[e.g.,][]{ginn2024glosslm}. 

\paragraph{}

We benchmark our approach on three different datasets, including the 2023 SIGMORPHON Shared Task data \cite{ginn-etal-2023-findings}, the GlossLM dataset \cite{ginn2024glosslm} that has the most extensive corpus of IGT available, and also FLORES \cite{goyal2022flores}; using state-of-the-art open-source instruction-tuned models, mainly Llama-3 \cite{Llama3modelcard} as well as Mixtral \cite{mixtral}. We find that \grammaMT{} can improve machine translation performance in low-resource setups, including endangered languages rarely encountered during pre-training. Even in high-resource languages, where the model has increased exposure and deeper understanding of the grammatical structure, we can observe substantial improvements from incorporating linguistic gloss resources into the prompt.

\section{Related work}
\paragraph{Machine translation with LLMs}has been extensively explored \citep{zhang-etal-2023-machine,garcia2023unreasonable,peng-etal-2023-towards,pourkamali2024machine}. Although LLMs perform well for high-resource languages they underperform for low-resource languages~\citep{hendy2023good,robinson-etal-2023-chatgpt,zhu2023multilingual}. While previous works study in-context learning for MT \cite{garcia2023unreasonable, puduppully2023decomposed, zhang2023prompting, sun2022zero}, effective alternatives that leverage linguistic information for unseen and low-resource languages remain underexplored.



\vspace{-0.1cm}

\paragraph{Using grammatical information with LLM}  Introducing grammatical information during training or inference can improve model performance~\citep{strubell-etal-2018-linguistically,cui2022lert,stahlberg-etal-2016-syntactically}. Similar to our work, \citet{zhou2020using} use glosses while training low-resource translation models. However, we use glosses in a training free approach and in the context of LLMs. \citet{tanzer2024a} and LingoLLM~\citep{zhang2024hire} use grammar books along with other resources to translate unseen and low-resource languages. Unlike these methods—which depend on grammar books, morphological analyzers, and dictionaries that are often unavailable—we use only a small number of gold or generated glosses, offering a more feasible solution for underrepresented languages. 



\section{\grammaMT ~}


We propose \grammaMT{}, a simple grammar-informed prompting approach for machine translation, wherein examples of Interlinear Gloss Texts (IGT) are used as a prompt to instruction-tuned LLMs. In doing so, our approach is essentially \textbf{training-free}. The approach also requires a \textbf{small set of support examples} and \textbf{minimal annotation time} (a handful of glosses by a linguistic or automatically generated by a model \cite{ginn2024glosslm}). In this section, 
we provide an overview of IGT and describe the proposed prompting of \grammaMT{}.




\paragraph{Interlinear Gloss Text Annotation.}
IGT annotations are triplets of source text, glosses for the source text, and fluent target translations for the source text.  The gloss consists of a sequence of target morphological annotations and (semantically full) lemmata for source words, indicating their grammatical morphemes and lexemes, shown by the following Swahili example. 

\texttt{\texttt{1. Source:} \textit{ (yeye) alimwona (yeye).}}

\texttt{2. Gloss:} {3SG -PST --see-FV 3SG}

\texttt{3. Translation:}  \textit{ S/he saw him/her}.
\\ 
In this example, the morphological annotation 
\texttt{3SG} stands for third-person singular and \texttt{PST} denotes the past tense of "see". Grammatical morphemes are labeled with uppercase letters. 
In contrast, lexemes (English lemma translations that convey semantic meaning) are labeled in lowercase (e.g., \emph{see}). In this way, IGT captures the syntax and morphology of a sentence, aiding to grasp the structure of the source language and to understand the relationship between input sentence and the translation. These glosses are the norm in linguistic descriptions
, and hence very common to find and easy to create. 

\paragraph{Prompting strategies.} \grammaMT{} augments an instruction-tuned LLM with in-context learning examples of interlinear glosses via three prompting strategies: \textbf{gloss-shot}, \textbf{chain-shot} and \textbf{model-gloss}, as illustrated in Figure \ref{fig:approach}. 

In the first prompting strategy, \textbf{gloss-shot}, the LLM is prompted to generate the translation \textbf{y} for the input sentence  \textbf{x} based on a set of $N$ interlinear-glossed text exemplars \textbf{g} (i.e., triples of source sentence, gloss line, translation), essentially predicting 
\( (\mathbf{g}_1, \cdots, \mathbf{g}_N, \mathbf{x})  \rightarrow \mathbf{y}\).

In the second prompting strategy, \textbf{chain-gloss}, the LLM is also conditioned on a set of $N$ interlinear-glossed text exemplars \textbf{g} to generate the translation, but in this strategy, the model first produces the gloss \( \mathbf{y}_{g} \) before formulating the translation \textbf{y}, essentially \( (\mathbf{g}_1, \cdots, \mathbf{g}_N, \mathbf{x} ) \rightarrow (\mathbf{y}_{g}, \mathbf{y})\). This prompting strategy can offer some insights into how the LLM arrived at a specific translation.

In the \textbf{model-gloss} strategy, a specialised gloss generation model (e.g., GlossLM \cite{ginn2024glosslm}) provides the gloss for the source sentence, rather than relying on the LLM to generate it itself. As with the other strategies, this one also includes in-context examples of interlinear-glossed text, followed by the source sentence. However, here the source sentence is paired with a gloss predicted by the external model \( \mathbf{y}_{ge} \), before the LLM produces the final translation: \( (\mathbf{g}_1, \cdots, \mathbf{g}_N, \mathbf{x}, \mathbf{y}_{ge} ) \rightarrow  \mathbf{y} \)

We illustrate the format of the prompt in Figure \ref{fig:approach} and in more detail in Appendix \ref{app:prompt}.

\section{Experimental setup}
\subsection{LLMs} We assess our \grammaMT{} approach using \texttt{Meta-Llama-3-70B-Instruct}~\cite{Llama3modelcard}, the recent instruction-tuned Llama with 70B parameters. Our machine translation approach does not involve any training. The translations are generated at inference time using a single A100 80GB GPU. We also report experiments with the smaller \texttt{Meta-Llama-3-8B-Instruct}, and \texttt{Mixtral-8x22B-Instruct-v0.1} \cite{mixtral}, as well as the closed-source GPT-4o model \cite{openai2024gpt4ocard} in Appendix \ref{app:model}. The open-source LLMs were loaded via the HuggingFace Hub library~\cite{wolf2020transformers} using 4-bit quantization, while the GPT-4o model was accessed through the OpenAI API\footnote{\url{https://platform.openai.com/}}. During inference, the models generate a translation using greedy decoding with a default temperature setting of 1.


\subsection{Prompting strategies and baselines}

\noindent \textbf{Baselines.} We first compare \grammaMT{} against other established in-context learning strategies, which use no explicit grammatical information:  
\begin{itemize}
    \item  \textbf{zero-shot:} Translation from the source to the target language without examples.
    \item \textbf{zero-CoT:} The LLM is prompted to think step by step before translating, again without examples.
    \item \textbf{few-shot:} The LLM translates the input using a few source-target example pairs.
\end{itemize}

We select zero-CoT over Chain-of-Thought, because our data lacked the detailed steps needed for MT breakdown. We also compare \grammaMT{} to the training-free LingoLLM \cite{zhang2024hire}, which uses more linguistic resources, including a grammar book, morphological analyzer, and a dictionary. For a thorough evaluation, we report performance of a state-of-the-art MT model, NLLB-200 (\texttt{nllb-200-distilled-600M}), while emphasising that it is not an LLM, as our focus is on improving LLMs for MT. Finally, we compare against a parallel dictionary baseline in Appendix \ref{app:other_baselines}.
\vspace{0.3cm}

\noindent\textbf{\grammaMT{} prompting.}  Our own approach augments few-shot prompting with grammatical information, where we explore three novel variants:
\begin{itemize}
    \item \textbf{\emph{gloss-shot}:} The LLM predicts based on examples that pair the source sentences not just with their translation but also with their gloss. 
    \item \textbf{\emph{chain-gloss}:} As in gloss-shot, but the LLM is additionally prompted to generate the gloss for the input sentence before translating. 
    \item \textbf{\emph{model-gloss}:} As in chain-gloss, but the gloss of the source sentence is obtained from an external gloss generation model and not from the LLM itself. For this, we use GlossLM \cite{ginn2024glosslm} that was trained to generate glosses.\footnote{See Appendix \ref{Appendix:glosslm_details} for details.}
\end{itemize}

For all prompting strategies\footnote{Except zero-shot and zero-CoT that have no examples.}, we use the same 21 translation examples per language, identified as the optimal value in our ablation studies (see Section \ref{sec:ablations-n}). Prompt templates are provided in Appendix \ref{app:prompt}. 



\subsection{Datasets and Languages}
\label{sec:data}

We evaluate translation quality across three datasets,  involving endangered, low-resource, and mid-to-high-resource languages, with English as the target language. \Cref{tab:languages} summarises the languages, scripts and test set sizes. For completeness, we also evaluate the reverse translation direction, with English as source language, in Section \ref{sec:ablations-n}.
 \begin{table}[!t]
\centering
\resizebox{0.5\textwidth}{!}{
\begin{tabular}{llllr}
\toprule
\textbf{Language}  & \textbf{Abbr.} & \textbf{Script}  & \textbf{Test}  & \textbf{Speakers}\\
\midrule
\multicolumn{5}{c}{\textbf{Sigmorphon dataset}}\\
\midrule
Gitksan    & Git           & Latin  &   37   & 1,110  \\
Lezgi      & Lez           & Cyrillic  &   87   & 800K \\
Natugu     & Ntu           & Latin  &   99   & 5,900  \\
Tsez       & Ddo           & Cyrillic  &   445   & 18K \\
\midrule
\multicolumn{5}{c}{\textbf{GlossLM dataset}}\\
\midrule
Swahili     & Swa          & Latin                & 439       & 200M \\
Yoruba      & Yor          & Latin w/ diac.     & 135       & 47M \\
Icelandic   & Ice          & Latin                & 27        & 330K \\
Marathi     & Mar          & Devanagari           & 43        & 83M \\
Kannada     & Kan          & Kannada              & 388       & 59M \\
Urdu        & Urd          & Perso-Arabic         & 259       & 232M \\
Thai        & Tha          & Thai                 & 352       & 61M \\
Greek       & Gre          & Greek                & 59        & 13.5M \\
Portuguese  & Por          & Latin                & 309       & 264M \\
Japanese    & Jap          & Japanese \footnotemark[3]     & 4,748     & 123M \\
Russian     & Rus          & Cyrillic             & 2,444     & 255M \\
Arabic      & Ara          & Arabic               & 136       & 274M \\
\bottomrule
\end{tabular}}
\caption{Overview of the languages and the test split sizes used in \grammaMT{} evaluation.}
\label{tab:languages}
\end{table}

\begin{table*}[!t]
\centering
\resizebox{\textwidth}{!}{
\begin{tabular}{cccccaccccaa}
\toprule
\multicolumn{1}{l}{{\textbf{Method}}}& \multicolumn{5}{c}{\textbf{BLEU}} & \multicolumn{5}{c}{\textbf{chrF++}} & \multicolumn{1}{c}{{\textbf{xCOMET}}}\\ \cmidrule{2-6} \cmidrule(l){7-11} \cmidrule(l){12-12} 
  & \textbf{Git} & \textbf{Lez} & \textbf{Ntu} & \multicolumn{1}{r|}{\textbf{Ddo}} & Avg.  & \multicolumn{1}{|r}{\textbf{Git}} & \textbf{Lez} & \textbf{Ntu} & \multicolumn{1}{r|}{\textbf{Ddo}} & \multicolumn{1}{a|}{Avg.} & Avg. \\
 \midrule
\multicolumn{1}{l|}{\text{NLLB-200}} & 0.9 & 0.8 & 0.4 & \multicolumn{1}{r|}{0.1} & 0.55 & \multicolumn{1}{|r}{23.65} & 18 & 12.3 & \multicolumn{1}{r|}{ 10.10} & \multicolumn{1}{a|}{13.80} & 12.82\\ 
\multicolumn{1}{l|}{\text{LingoLLM w/ GPT-4}} & \underline{14.3} & - & \underline{12.9} & \multicolumn{1}{r|}{\textbf{15.1}} & \underline{14.1} &\multicolumn{1}{|r}{-} & - & - & \multicolumn{1}{r|}{-} & \multicolumn{1}{a|}{-} & - \\
\hline\hline

\multicolumn{1}{l|}{\text{zero-shot}} &1.26 & 1.46 & 0.26 & \multicolumn{1}{r|}{0.39} & 0.88 & \multicolumn{1}{|r}{23.90} & 17.71 & 13.76 & \multicolumn{1}{r|}{16.84} & \multicolumn{1}{a|}{18.05} & 15.21 \\ 
\multicolumn{1}{l|}{\text{zeroCoT}} & 2.84    & 1.74 &    0.37 & \multicolumn{1}{r|}{    0.32} & 1.32 & \multicolumn{1}{|r}{21.21} & 15.27     & 13.95 & \multicolumn{1}{r|}{15.68} & \multicolumn{1}{a|}{16.53} & 14.50 \\ 
\multicolumn{1}{l|}{\text{few-shot}} & 4.71 & 6.36 & 3.34 & \multicolumn{1}{r|}{1.46} & 3.94 &\multicolumn{1}{|r}{25.18} & 22.89 & 19.41 & \multicolumn{1}{r|}{20.03} & \multicolumn{1}{a|}{21.85} & 16.76\\
\hline
\hline
\multicolumn{1}{l|}{\text{gloss-shot}} & 4.96 & 5.80 & 1.32 & \multicolumn{1}{r|}{1.72} & 3.41 &\multicolumn{1}{|r}{\underline{25.87}} & \underline{23.08} & \underline{20.24} & \multicolumn{1}{r|}{\underline{20.95}} & \multicolumn{1}{a|}{\underline{22.50}} & \underline{18.21}\\
\multicolumn{1}{l|}{\text{chain-gloss}} & 5.71 & \underline{7.29} & 2.35 & \multicolumn{1}{r|}{1.63} &4.25 & \multicolumn{1}{|r}{24.66} & 22.62 & 19.19 & \multicolumn{1}{r|}{18.01} & \multicolumn{1}{a|}{20.84} & 16.78\\ 
\multicolumn{1}{l|}{\text{model-gloss}} &\textbf{18.7} &   \textbf{13.94}& \textbf{16.96}&\multicolumn{1}{r|}{\underline{14.28}} &\textbf{15.97} & \multicolumn{1}{|r}{\textbf{47.89}} & \textbf{39.65} & \textbf{41.56} & \multicolumn{1}{r|}{\textbf{42.30}} & \multicolumn{1}{a|}{\textbf{41.45}} &  \textbf{40.83}\\ 
\bottomrule
\end{tabular}
}
\caption{\grammaMT{}'s performance (using Llama-3 70B)  for \textbf{unseen/endangered languages} on the 2023 SIGMORPHON test split, against in-context baselines and SOTA models like NLLB-200 and LingoLLM. Best results are in bold and second-best are underlined.}
\label{tab:unseen}
\end{table*}

\begin{table*}[h!]
\centering
\resizebox{\textwidth}{!}{
\begin{tabular}{ccccccacccccaa}
\toprule
\multicolumn{1}{l}{{\textbf{Method}}}& \multicolumn{6}{c}{\textbf{BLEU}} & \multicolumn{6}{c}{\textbf{chrF++}} & \multicolumn{1}{c}{\textbf{xC}}\\ \cmidrule{2-7} \cmidrule(l){8-13} \cmidrule(l){14-14}
 & \textbf{Swa} & \textbf{Yor} & \textbf{Ice} & \textbf{Mar} & \multicolumn{1}{r|}{\textbf{Kan}}  & Avg.  & \multicolumn{1}{|r}{\textbf{Swa}} & \textbf{Yor} & \textbf{Ice} & \textbf{Mar} & \multicolumn{1}{r|}{\textbf{Kan}}  & Avg. & \multicolumn{1}{|a}{Avg.}\\
\midrule

 \multicolumn{1}{l|}{\text{NLLB-200}} & 6.9 & 0.5 & 3.5     & 0.3 &  \multicolumn{1}{r|}{0.8    } & 2.4 & \multicolumn{1}{|r}{24.2} & 10.8 & 21.1 & 10     & \multicolumn{1}{r|}{ 10.7    } & 15.36 & \multicolumn{1}{|a}{20.21} \\ 
\hline\hline

 \multicolumn{1}{l|}{\text{zero-shot}} & 16.99 & 4.48 & 4.92 & 0.70 &  \multicolumn{1}{r|}{5.84} & 6.58 & \multicolumn{1}{|r}{40.35} & 18.87 & 27.97 & 13.28 & \multicolumn{1}{r|}{ 25.65} & 25.22 & \multicolumn{1}{|a}{27.10}\\ 
\multicolumn{1}{l|}{\text{zero-CoT}} & 15.78 & 1.93 &  4.64 & 1.08 & \multicolumn{1}{r|}{ 4.99 } & 5.69 & \multicolumn{1}{|r}{39.15} & 18.84 &  \underline{28.02} & 14.87 & \multicolumn{1}{r|}{25.20} &  25.22 & \multicolumn{1}{|a}{27.76} \\

\multicolumn{1}{l|}{\text{few-shot}} & \underline{22.41} & 11.98 &  \textbf{6.43} & \textbf{19.19} & \multicolumn{1}{r|}{ \underline{23.50} } & \underline{16.69} & \multicolumn{1}{|r}{\underline{45.75}} & 29.92 &  \textbf{28.87} & \underline{36.11} & \multicolumn{1}{r|}{\underline{44.16}} &  \underline{36.96} & \multicolumn{1}{|a}{34.52} \\
\hline\hline

\multicolumn{1}{l|}{\text{gloss-shot}} & 22.18 & \textbf{16.32} & 3.50 & \underline{17.53} &\multicolumn{1}{r|}{ 22.35 } & 16.39  & \multicolumn{1}{|r}{\textbf{46.50}} & \underline{33.24} &  25.79 & \textbf{36.18} & \multicolumn{1}{r|}{42.68} & 36.88 & \multicolumn{1}{|a}{\underline{35.65}}  \\
\multicolumn{1}{l|}{\text{chain-gloss}} &  \textbf{23.53} & \underline{14.10} & \underline{5.05} & 17.32 &  \multicolumn{1}{r|}{\textbf{25.25}} & \textbf{17.06}  & \multicolumn{1}{|r}{45.44} & \textbf{33.54} & 24.90 & 35.37 & \multicolumn{1}{r|}{\textbf{46.27}} & \textbf{37.10} & \multicolumn{1}{|a}{\textbf{35.77}} \\ 

\bottomrule
\end{tabular}
}
\caption{\grammaMT{}'s performance (using Llama-3 70B) for \textbf{low-resource languages} on the GlossLM data, the largest corpus of IGT data. Best results are in bold; second-best underlined.  xC is xCOMET.}
\label{tab:low-resource}
\end{table*}

\paragraph{Sigmorphon:} We use the dataset from the 2023 SIGMORPHON Shared Task for evaluating on unseen, endangered languages \cite{ginn-etal-2023-findings}, with Gitksan, Lezgi, Natugu, and Tsez. This dataset includes translation pairs from each source language to English, together with the interlinear glosses and morphological segmentation of the source sentences. 
We report performance on the test set, while the validation split is used for ablation studies. In both cases, support examples are drawn from the training split, specifically the first 21 sentences (Section \ref{sec:ablations-n} shows that $N = 21$ is optimal).

\paragraph{GlossLM corpus:} For evaluating on low to high-resource languages, we use the GlossLM dataset \cite{ginn2024glosslm}, a recent and extensive compilation of interlinear glossed text (IGT) from six different IGT corpora. This dataset includes 250k unique sentences across 1800 languages. We selected languages from different scripts, specifically considering Swahili, Yoruba, Icelandic, Marathi, and Kannada for low-resource languages. For mid-to-high-resource languages, we included Urdu, Thai, Greek, Portuguese, Japanese, Russian, and Arabic. However, the GlossLM dataset only provides evaluation splits (dev/test) for the endangered languages included in the SIGMORPHON Shared Task, as this data is the most consistent. For other languages ranging from low to high-resource, the dataset offers only a training split. To address this, we created evaluation splits by designating most of the training set for testing, reserving the first 21 examples for in-context learning (Section \ref{sec:ablations-n} provides empirical evidence that $N = 21$ is optimal). We have detailed the number of test samples for each language in \Cref{tab:languages}. To avoid unfair evaluation, results for the model-gloss strategy are not provided on our test split, 
since the GlossLM model \cite{ginn2024glosslm} used in this strategy was exposed to those training samples. But we report model-gloss results for these languages in the subsequent dataset.

\paragraph{FLORES-200:} We also report results on the FLORES dataset \cite{goyal2022flores} (test split). We use the same languages we considered from the GlossLM dataset, and the same set of 21 examples since FLORES does not contain the annotated glosses, to assess our approach's ability to generalise in the absence of in-domain glosses.

\subsection{Metrics}
For evaluation, we report MT evaluation metrics, namely BLEU \cite{papineni2002bleu} with SacreBLEU tokenisation \cite{post2018call},  and the chrF++ metric, which exhibits a stronger correlation with human scores \cite{popovic2017chrf++}. To further strengthen our evaluations, we include a model-based metric using xCOMET-XXL \cite{10.1162/tacl_a_00683}, the latest version of the widely adopted COMET model~\cite{rei-etal-2020-unbabels}. We report significance tests over these metrics in Appendix \ref{app:sig}.


\section{Results}
\label{sec:results}

\begin{table*}[!t]
\centering
\resizebox{\textwidth}{!}{
\begin{tabular}{ccccccccacccccccaa}
\toprule
\multicolumn{1}{l}{\textbf{Method}} & \multicolumn{8}{c}{\textbf{BLEU}} & \multicolumn{8}{c}{\textbf{chrF++}} & \multicolumn{1}{c}{\textbf{xC}}\\ \cmidrule{2-9} \cmidrule(l){10-17} \cmidrule(l){18-18}
 & \textbf{Urd} & \textbf{Tha} & \textbf{Gre} & \textbf{Por} & \textbf{Jap} & \textbf{Rus} & \multicolumn{1}{r|}{\textbf{Ara}} & Avg.  &  \multicolumn{1}{|r}{\textbf{Urd}} & \textbf{Tha} & \textbf{Gre} & \textbf{Por} & \textbf{Jap} & \textbf{Rus} & \multicolumn{1}{r|}{\textbf{Ara}} & Avg. & \multicolumn{1}{|a}{Avg.}\\
\midrule

\multicolumn{1}{l|}{NLLB-200} & 0.2     & 0.2 & 0.5     & 26.2 & 0.4    & 2.4 & \multicolumn{1}{r|}{1.4 } & 4.47 & \multicolumn{1}{|r}{9.1} & 9.3     & 11.3 & 47 & 14.4 & 17     & \multicolumn{1}{r|}{11.5} & 17.09 & \multicolumn{1}{|a}{24.25}\\
\hline\hline
\multicolumn{1}{l|}{\text{zero-shot}} & 4.00 & 1.35 & 6.13 & 37.75 & 7.17 & \underline{25.12} & \multicolumn{1}{r|}{3.46 } & 12.15 & \multicolumn{1}{|r}{20.53} & 12.56 & 23.14 & 59.21 & 27.62 & 47.31 & \multicolumn{1}{r|}{19.95} & 30.05 & \multicolumn{1}{|a}{35.42} \\

\multicolumn{1}{l|}{\text{zero-CoT}} & 4.71 & 1.80     & 8.22 & 37.20     & 7.26  & 23.42 & \multicolumn{1}{r|}{3.81 } & 12.35 & \multicolumn{1}{|r}{22.60} & 12.91 & 25.56 & 56.50 & 27.30  & 45.07 & \multicolumn{1}{r|}{19.18} & 29.87 & \multicolumn{1}{|a}{35.10} \\

\multicolumn{1}{l|}{\text{few-shot}} & 26.19 & \underline{7.68} & \underline{10.62} & 4\underline{4.14} & \underline{13.74} & 24.94 & \multicolumn{1}{r|}{\underline{5.35}}  & \underline{18.95} & \multicolumn{1}{|r}{43.36} & \underline{19.76} & \textbf{27.55} & \textbf{63.88} & \underline{35.94} & \underline{48.59} & \multicolumn{1}{r|}{\textbf{21.28}} & \underline{37.19} & \multicolumn{1}{|a}{41.03} \\
\hline\hline
\multicolumn{1}{l|}{\text{gloss-shot}} & \underline{26.86} & 6.26 & 9.56 & \textbf{44.37} & 13.65 & 23.99 & \multicolumn{1}{r|}{ \textbf{5.60}}  & 18.61 & \multicolumn{1}{|r}{\underline{43.49}} & 19.27 & \underline{27.17} & \underline{63.72} & 35.71 & 48.13 & \multicolumn{1}{r|}{\underline{21.19}} & 36.95 & \multicolumn{1}{|a}{\underline{41.05}} \\

\multicolumn{1}{l|}{\text{chain-gloss}} & \textbf{28.71} & \textbf{8.34} & \textbf{10.74 }& 42.88 & \textbf{15.41} & \textbf{27.92} & \multicolumn{1}{r|}{5.26} & \textbf{19.75} & \multicolumn{1}{|r}{\textbf{45.86}} & \textbf{19.81} & 27.11 & 62.33 & \textbf{37.29} & \textbf{50.22} & \multicolumn{1}{r|}{19.51} &  \textbf{37.20} & \multicolumn{1}{|a}{\textbf{41.46}} \\



\bottomrule
\end{tabular}
}
\caption{\grammaMT{}'s performance (using Llama-3 70B) for \textbf{mid-high-resource languages}  on the GlossLM data. Best results are in bold; second-best underlined.  xC is xCOMET.}
\label{tab:mid-high}
\end{table*}

\paragraph{\grammaMT{} outperforms in unseen/endangered languages.}

In Table ~\ref{tab:unseen}, we show how \grammaMT{} performs on four endangered languages: Gitksan, Lezgi, Natugu and Tsez (all unseen by the LLM during pre-training).
The results demonstrate that the model-gloss strategy consistently outperforms the baselines across the three metrics. Focusing on BLEU, this strategy shows a large improvement of 15.09, 14.65, and 12.03 BLEU points against zero-shot, zero-CoT and the few-shot approach on average, respectively. Additionally, it surpasses the specialised NLLB translation model, which struggles with unseen languages. Furthermore, the model-gloss strategy outperforms \textsc{LingoLLM} \cite{zhang2024hire}, the state-of-the-art training-free method in this shared task, by over 4 BLEU points for Gitksan and Lezgi, while being only slightly outperformed by 0.82 points for Tsez. This is despite LingoLLM's leveraging vastly more extensive linguistic resources, such as grammar books and dictionaries.

Within the \grammaMT{} strategies, model-gloss is more robust compared to relying on the LLM for gloss prediction (chain-gloss)\footnote{See Section~\ref{sec:gloss_performance} for a comparison of gloss performance.} or using glosses only for examples (gloss-shot). This is most likely because it relies on a specialised gloss model tailored to these languages. However, both these methods still show promising results. We see that the gloss-shot strategy outperforms the prompting baselines across all unseen languages tested on using the chrF++ metric. Additionally, BLEU scores improve for both Gitksan and Tsez. For chain-gloss, while few-shot outperforms with the chrF++ metric, we observe BLEU score increases of 1 point for Gitksan, 0.93 for Lezgi, and 0.17 for Tsez. Overall, ~\grammaMT{} outperforms translation for unseen languages in our experiments, indicating the benefits in this challenging language setup.
\paragraph{Chain-gloss improves translation of low-resource languages.}

We also assess \grammaMT{} on low-resource languages, including Swahili, Yoruba, Icelandic, Marathi and Kannnada (see Table ~\ref{tab:low-resource}). Chain-gloss improves the performance on the majority of them as seen in the average BLEU, chrF++ and the xCOMET score. This improvement is similarly observed with gloss-shot, particularly in the chrF++ performance for Swahili and Marathi. Notably, we observed a large improvement for Yoruba from adding the gloss to the context, with an increase of more than 4 BLEU points and 3 chrF++ points compared to few-shot. Icelandic and Marathi, exhibited the best performance using few-shot based on BLEU. We exclude the model-gloss strategy, as it leverages glosses from GlossLM~\cite{ginn2024glosslm}. As GlossLM was pre-trained on this data, including the model-gloss strategy would lead to unfair evaluation, due to prior exposure to the test set.

\paragraph{Chain-gloss also improves mid-high-resource languages.}



In Table~\ref{tab:mid-high}, we observe that \grammaMT{} improves the performance for all of the high-resource languages on BLEU, with the best performing method being either chain-gloss or gloss-shot. 
Notably, Urdu and Russian show substantial improvements, with chain-gloss surpassing few-shot by more than 2.5 BLEU points. Using chrF++, consistent with the BLEU results, we have chain-gloss outperforming the other methods except for Portuguese, Arabic and Greek, for which few-shot outperforms both gloss-shot and chain-gloss. For these languages, gloss-shot also outperforms chain-gloss. We again excluded results for the model-input strategy as the gloss model had prior exposure to the test set. Overall, results show that augmenting the context with grammatical information is not only beneficial in low-resource settings, 
but also for mid-to-high-resource languages.

\subsection{Out of domain evaluation: Flores}

\begin{table*}[!h]
\centering
\resizebox{\textwidth}{!}{
\begin{tabular}{rrlrrrrrrrrrra}
\toprule
\multicolumn{1}{l}{\textbf{Method}}&  \multicolumn{13}{c}{\textbf{BLEU}}\\
\cmidrule{2-14}
\multicolumn{1}{l}{~} &  \multicolumn{1}{c}{\textbf{Swa}} & \multicolumn{1}{c}{\textbf{Yor}} & \multicolumn{1}{c}{\textbf{Ice}}   & \multicolumn{1}{c}{\textbf{Mar}} & \multicolumn{1}{c|}{\textbf{Kan}} & \multicolumn{1}{c}{\textbf{Urd}} & \multicolumn{1}{c}{\textbf{Tha}} & \multicolumn{1}{c}{\textbf{Gre}} & \multicolumn{1}{c}{\textbf{Por}} & \multicolumn{1}{c}{\textbf{Jap}} & \multicolumn{1}{c}{\textbf{Rus}} & \multicolumn{1}{c}{\textbf{Ara}} & \multicolumn{1}{|a}{\textbf{Avg.}}\\ \midrule 

\multicolumn{1}{l|}{few-shot} & \underline{20.99} & 3.94 & \underline{18.23} & \textbf{18.72} & \multicolumn{1}{r|}{\underline{3.63}} & \textbf{19.78} & \textbf{21.34} & 28.07 & 41.24 & 16.62 & 27.27 & \multicolumn{1}{r|}{\textbf{28.59}} & 20.69 \\ 

\multicolumn{1}{l|}{gloss-shot} &  \textbf{22.37} & \underline{5.00} & \textbf{19.40} & \underline{18.26} & \multicolumn{1}{r|}{3.04}  & 18.65 & \underline{20.35} & \textbf{30.08} & \underline{43.30} & \underline{19.61}  & \underline{31.14} & \multicolumn{1}{r|}{\underline{28.43}} & \textbf{21.64}\\ 

\multicolumn{1}{l|}{chain-gloss} & 20.26 & \textbf{5.03} & 18.05 & 17.20 & \multicolumn{1}{r|}{\textbf{4.40}} & 18.13 & 19.32 & 27.82 & 41.62 & 18.23 & 30.26 & \multicolumn{1}{r|}{26.74} &  20.59\\ 

\multicolumn{1}{l|}{model-gloss} & 18.30 & 3.67& 16.92  & 17.73 & \multicolumn{1}{r|}{3.09} & \underline{18.75} & 20.21 & \underline{29.15} & \textbf{43.64}  &  \textbf{19.77} & \textbf{31.16} & \multicolumn{1}{r|}{\textbf{28.59}} & \underline{20.92}  \\ 
\bottomrule
\end{tabular}
}
\caption{BLEU performance on the FLORES test set. We select the 21-shot examples from the GlossLM data, as FLORES lacks annotated glosses. Results show that \grammaMT{} can generalise in an out-of-domain setting.}
\label{tab:flores_business}
\end{table*}

\begin{table*}[htbp!]
\centering
\resizebox{\textwidth}{!}{
\begin{tabular}{ll|ccccacccca}
\toprule
\multicolumn{1}{l}{{\textbf{Method}}}&   \multicolumn{1}{c}{{\textbf{Model}}} & \multicolumn{5}{c}{\textbf{BLEU}} & \multicolumn{5}{c}{\textbf{chrF++}} \\ \cmidrule{3-7} \cmidrule(l){8-12}
  & \multicolumn{1}{c}{~}& \multicolumn{1}{c}{\textbf{Git}} & \textbf{Lez} & \textbf{Ntu} & \multicolumn{1}{r|}{\textbf{Ddo}} & Avg.  & \multicolumn{1}{|r}{\textbf{Git}} & \textbf{Lez} & \textbf{Ntu} & \multicolumn{1}{r|}{\textbf{Ddo}} & Avg. \\
 \midrule

\multicolumn{1}{l|}{\text{few-shot}} & Llama-3 70B & 4.71 & 6.36 & \underline{3.34} & \multicolumn{1}{r|}{1.46} & 3.94 &\multicolumn{1}{|r}{25.18} & 22.89 & 19.41 & \multicolumn{1}{r|}{20.03} & 21.85 \\

\multicolumn{1}{l|}{\text{gloss-shot}} & Llama-3 70B &4.96 & 5.80 & 1.32 & \multicolumn{1}{r|}{1.72} & 3.41 &\multicolumn{1}{|r}{\underline{25.87}} & \underline{23.08} & \underline{20.24} & \multicolumn{1}{r|}{\underline{20.95}} & \underline{22.50}\\
\multicolumn{1}{l|}{\text{chain-gloss}}  & Llama-3 70B &\underline{5.71} & \underline{7.29} & 2.35 & \multicolumn{1}{r|}{\underline{1.63}} &\underline{4.25} & \multicolumn{1}{|r}{24.66} & 22.62 & 19.19 & \multicolumn{1}{r|}{18.01} & 20.84\\

\multicolumn{1}{l|}{\text{model-gloss}}  & Llama-3 70B &\textbf{18.7} &   \textbf{13.94}& \textbf{16.96}&\multicolumn{1}{r|}{\textbf{14.28}} &\textbf{15.97} & \multicolumn{1}{|r}{\textbf{47.89}} & \textbf{39.65} & \textbf{41.56} & \multicolumn{1}{r|}{\textbf{42.30}} & \textbf{41.45} \\ \hline

\multicolumn{1}{l|}{\text{few-shot}} & Llama-3 8B & 2.30 & 5.03 & 1.70 & \multicolumn{1}{r|}{0.47} & 2.38 &\multicolumn{1}{|r}{\underline{25.28}} & 20.4 & \underline{18.9} & \multicolumn{1}{r|}{18.64} & 20.81 \\

\multicolumn{1}{l|}{\text{gloss-shot}} & Llama-3 8B & 2.83 & 4.63 & 1.46 & \multicolumn{1}{r|}{0.60} & 2.38 & \multicolumn{1}{|r}{23.59} & 20.8 & 17.8 & \multicolumn{1}{r|}{\underline{18.8}} & 20.25\\

\multicolumn{1}{l|}{\text{chain-gloss}} & Llama-3 8B & \underline{4.21} & \underline{8.2} &\underline{2.60} & \multicolumn{1}{r|}{\underline{0.80}} & \underline{3.95} & \multicolumn{1}{|r}{23.26} & \underline{32.6} & 17.30 & \multicolumn{1}{r|}{16.81} & \underline{22.49}\\

\multicolumn{1}{l|}{\text{model-gloss}}  & Llama-3 8B &\textbf{7.11} & \textbf{10.68} &\textbf{7.44} & \multicolumn{1}{r|}{\textbf{8.2}} &\textbf{8.36} & \multicolumn{1}{|r}{\textbf{37.72}} & \textbf{34.41} & \textbf{32.80} & \multicolumn{1}{r|}{\textbf{35.09}} & \textbf{35.00}\\ \hline

\multicolumn{1}{l|}{\text{few-shot}} & Mixtral-8x22B & 3.32 & \underline{7.05} & 3.80 & \multicolumn{1}{r|}{2.46} & 4.16 &\multicolumn{1}{|r}{\underline{25.04}} & \underline{23.45} & 21.59 & \multicolumn{1}{r|}{20.76} & 22.71 \\

\multicolumn{1}{l|}{\text{gloss-shot}} & Mixtral-8x22B & \underline{4.58} & \underline{6.56} & \underline{4.63} & \multicolumn{1}{r|}{\underline{3.08}} & \underline{4.71} & \multicolumn{1}{|r}{24.80} & 22.45 & \underline{22.14} & \multicolumn{1}{r|}{ \underline{21.52}} & \underline{22.73}\\

\multicolumn{1}{l|}{\text{chain-gloss}} & Mixtral-8x22B & 3.12 & 1.50 &4.56 & \multicolumn{1}{r|}{1.06} & 2.56 & \multicolumn{1}{|r}{18.55} & 10.88 & 21.94 & \multicolumn{1}{r|}{15.28} & 16.66\\

\multicolumn{1}{l|}{\text{model-gloss}}  & Mixtral-8x22B &\textbf{16.27} & \textbf{13.74} &\textbf{19.24} & \multicolumn{1}{r|}{\textbf{18.89}} &\textbf{17.03} & \multicolumn{1}{|r}{\textbf{48.45}} & \textbf{40.45} & \textbf{44.71} & \multicolumn{1}{r|}{\textbf{44.87}} & \textbf{44.62} \\ \hline

\multicolumn{1}{l|}{\text{few-shot}} & GPT-4o & 5.49 & \underline{8.25} & 4.14 & \multicolumn{1}{r|}{1.64} & 4.88 &\multicolumn{1}{|r}{27.58} & \underline{23.32} & 21.50 & \multicolumn{1}{r|}{19.29} & 22.71 \\

\multicolumn{1}{l|}{\text{gloss-shot}} & GPT-4o & \underline{5.87} & 7.29 &  3.93 & \multicolumn{1}{r|}{\underline{2.04}} & 4.78 & \multicolumn{1}{|r}{\underline{28.49}} & 22.51 & 21.68 & \multicolumn{1}{r|}{ 19.77} & 22.73\\

\multicolumn{1}{l|}{\text{chain-gloss}} & GPT-4o &  4.16 & 7.46 & \underline{5.72} & \multicolumn{1}{r|}{1.91} & \underline{4.81} & \multicolumn{1}{|r}{26.23} & 23.15 &  \underline{21.77} & \multicolumn{1}{r|}{\underline{20.29}} & \underline{22.86} \\

\multicolumn{1}{l|}{\text{model-gloss}}  & GPT-4o &\textbf{22.24} & \textbf{ 14.45} &\textbf{21.25} & \multicolumn{1}{r|}{\textbf{ 17.10}} &\textbf{18.69} & \multicolumn{1}{|r}{\textbf{49.59}} & \textbf{38.64} & \textbf{45.03} & \multicolumn{1}{r|}{\textbf{41.37}} & \textbf{43.66} \\ 

\bottomrule

\end{tabular}
}
\caption{BLEU performance of \grammaMT{} on the 2023 SIGMORPHON test split across the different models (Llama-3 70B, Llama-3 8B, Mixtral-8x22B, GPT-4o).}
\label{tab:modelsize_unseen}
\end{table*}

We also evaluate \grammaMT{} on the FLORES test set, where in-domain glosses are unavailable, by reusing the same GlossLM examples in the translation prompts. \Cref{tab:flores_business} shows that gloss-shot achieves the highest average BLEU score, followed by model-gloss, with both achieving notable improvements of 2 points for Portuguese, Japanese, and Russian over few-shot. This suggests that both strategies can be effective even without annotated glosses for the current domain. In contrast, chain-gloss often struggles to predict accurate glosses and translations, likely due to a distributional shift from the short, simple GlossLM examples, to the more complex and lengthy input sentences in the FLORES dataset.  The example in Figure \ref{fig:flores} of Appendix \ref{flores:chrf} illustrates. The model-gloss strategy also performs poorly for low-resource languages. Thus, in out-of-domain settings, it is preferable to use glosses as examples (gloss-shot) rather than having the model generating the gloss without in-domain examples, to avoid misleading translations.



\paragraph{\grammaMT{} generalizes effectively across different LLM architectures and sizes.}


In addition to evaluating our approach using Llama-3 70B, we assess its ability to generalize to other models. Specifically, we report results for Llama-3 8B and Mixtral-8x22B, as well as the closed-source model GPT-4o. Table~\ref{tab:modelsize_unseen} shows the performance of these models for the unseen, endangered languages on the 2023 SIGMORPHON test split. Refer to Appendix \ref{app:model} for the results across the remaining languages. As shown in Table~\ref{tab:modelsize_unseen}, \grammaMT{} generalizes well to other LLMs, yielding a stronger performance with GPT-4o and Mixtral. The smaller Llama-3 8B model also benefits from incorporating grammatical information, with model-gloss and chain-gloss outperforming the few-shot baseline on average across both BLEU and chrF++. Overall, these results provide evidence that \grammaMT{} is a versatile approach that achieves good performance with both small and large models.

\begin{figure*}[t]
  \centering
  \includegraphics[width=\linewidth]{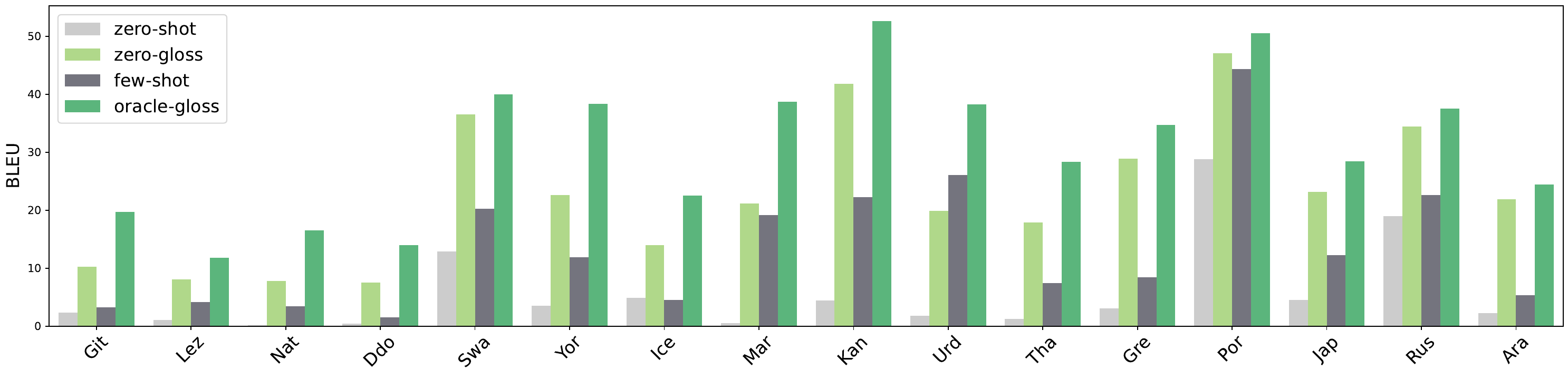}
  \caption{Simulation of an oracle experiment with \grammaMT{} using reference glosses (\emph{oracle-gloss} with $N$-shot examples or \emph{zero-gloss}) to assess if performance improves with accurate generation or access to correct glosses.}
  \label{fig:zeros}
\end{figure*}

\begin{figure}[htbp!]
  \centering
    \includegraphics[width=\linewidth]{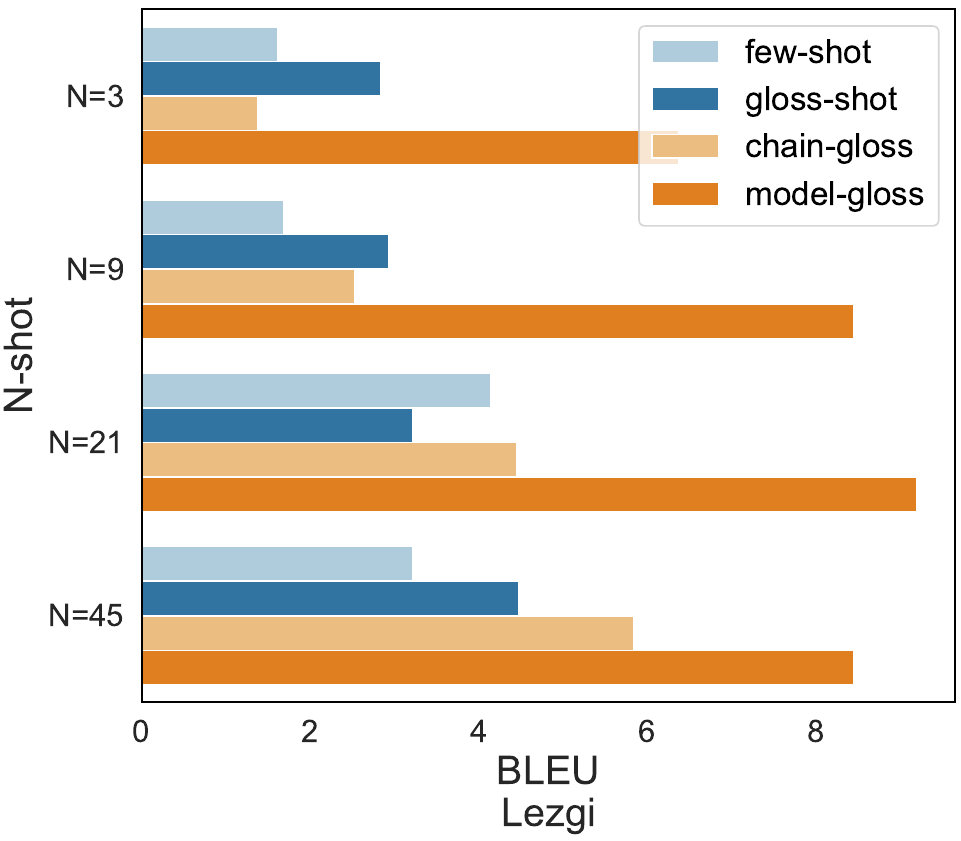}
  \caption{Varying $N$-shot examples from 3 to 45. These ablations were conducted on the validation split of the 2023 SIGMORPHON Shared Task data for Lezgi.}
  \label{fig:n-shot}
\end{figure}
\section{Further analysis and discussion}
\label{sec:ablations-n}

We conduct a series of ablation studies on the validation splits of the aforementioned datasets to better understand the impact of \grammaMT{} on improving LLM performance in machine translation.

\paragraph{Varying $N$.}



We consider the impact of the number of examples provided in prompts and vary the number of shots, $N$, both in our proposed \grammaMT{} strategy and in the few-shot baseline. We illustrate this for Lezgi in \Cref{fig:n-shot}. An increase of $N$ leads to improvements in all strategies, with optimal value being $N=21$.  We see large gains on chain-gloss by increasing $N$, suggesting that chain-gloss needs a sufficient number of examples to demonstrate the process of generating glosses.


\paragraph{Gloss Accuracy.}
\label{sec:gloss_performance}

\begin{figure}[h]
  \centering
    \includegraphics[width=\linewidth]{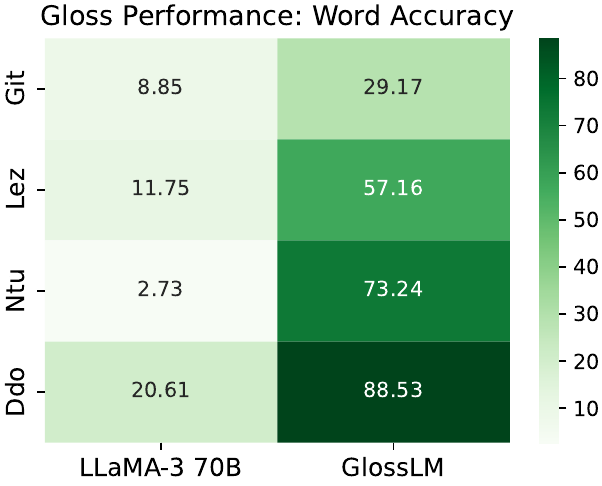}
  \caption{Evaluation of glosses generated by chain-gloss (Llama-3 70B) and model-gloss (GlossLM).}
  \label{fig:gloss_perf}
\end{figure}

Here we study to what extent do the glosses generated by chain-gloss and model-gloss strategies influence the translation output. 
We compare the glosses generated by Llama (used in the chain-gloss strategy) with GlossLM \cite{ginn2024glosslm} (used in the model-gloss strategy). Figure \ref{fig:gloss_perf} highlights that Llama struggles with gloss accuracy for rarely seen languages, achieving less than 21\% accuracy for Tsez (Ddo). In contrast, GlossLM performs substantially better, achieving up to 88\% for Tsez, directly contributing to the model-gloss strategy’s superior MT performance in Table \ref{tab:unseen}. We used word accuracy to assess gloss performance, consistent with the evaluation in GlossLM's work \cite{ginn2024glosslm}, reporting further metrics in Appendix \ref{app:gloss_perf}.

\paragraph{Oracle Setup.}


Here we further study translation performance if the model could accurately generate or access the gloss of the source sentence. We conduct an oracle experiment where we replace the generated glosses in the chain-gloss or model-gloss strategies with gold-standard glosses (\emph{oracle-gloss}). We also evaluate a zero-shot setup (\emph{zero-gloss}), prompting the model to translate directly from the source with the gold gloss. Both are compared to their respective baselines (few-shot and zero-shot).


%

Oracle-gloss significantly improves by an average of 17.46 BLEU points ($\pm$ 6.6) over few-shot across all languages, and zero-gloss also outperforms zero-shot by a massive margin of 16.02 BLEU points ($\pm$ 8.89). Notably, zero-gloss even surpasses the few-shot setting that uses machine translation examples. Overall, these results highlight the potential of leveraging glosses for improving machine translation. A promising direction is the development of automatic gloss models, such as GlossLM \cite{ginn2024glosslm}.

\begin{figure}[htbp!]
  \centering
    \includegraphics[width=\linewidth]{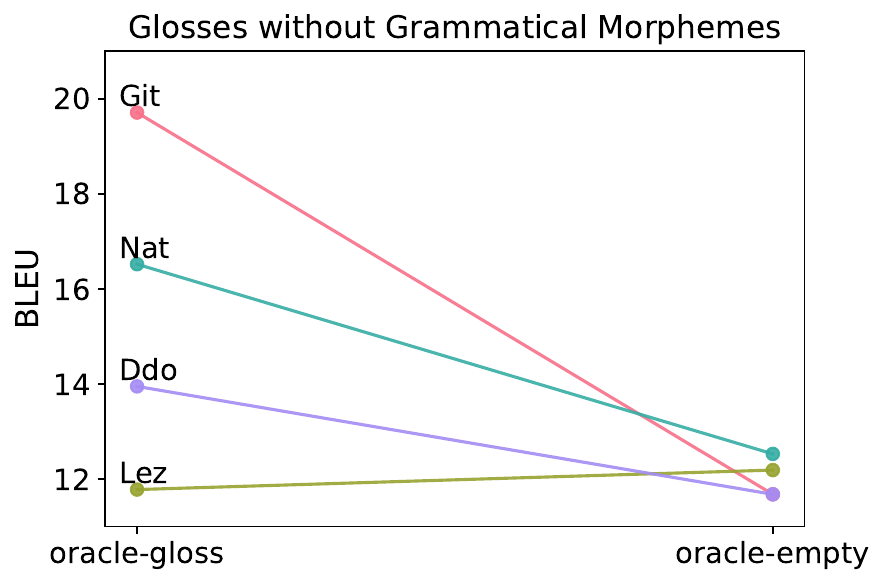}
  \caption{Performance drops when removing grammatical annotations (oracle-empty) compared to the original glosses (oracle-gloss). These ablations were also conducted on the validation split of the SIGMORPHON.}
  \label{fig:remove_gramma}
\end{figure}

\paragraph{Role of grammatical annotations.}

We further analyze whether performance is solely due to the English lemmata or whether grammatical annotations actually matter. Figure \ref{fig:remove_gramma} shows a performance drop when grammatical labels are removed, indicating their importance beyond mere word-by-word translation from lemmata. Moreover, we also present examples of translations produced by \grammaMT~ in Appendix \ref{app:examples} where we further observe that our strategies generate more satisfactory translations compared to the few-shot approach by being grammatical-aware. In Appendix \ref{app:seg}, we also explore other grammatical augmentations.

\paragraph{MT from English (en →).}

\begin{figure}[htbp!]
  \centering
    \includegraphics[width=\linewidth]{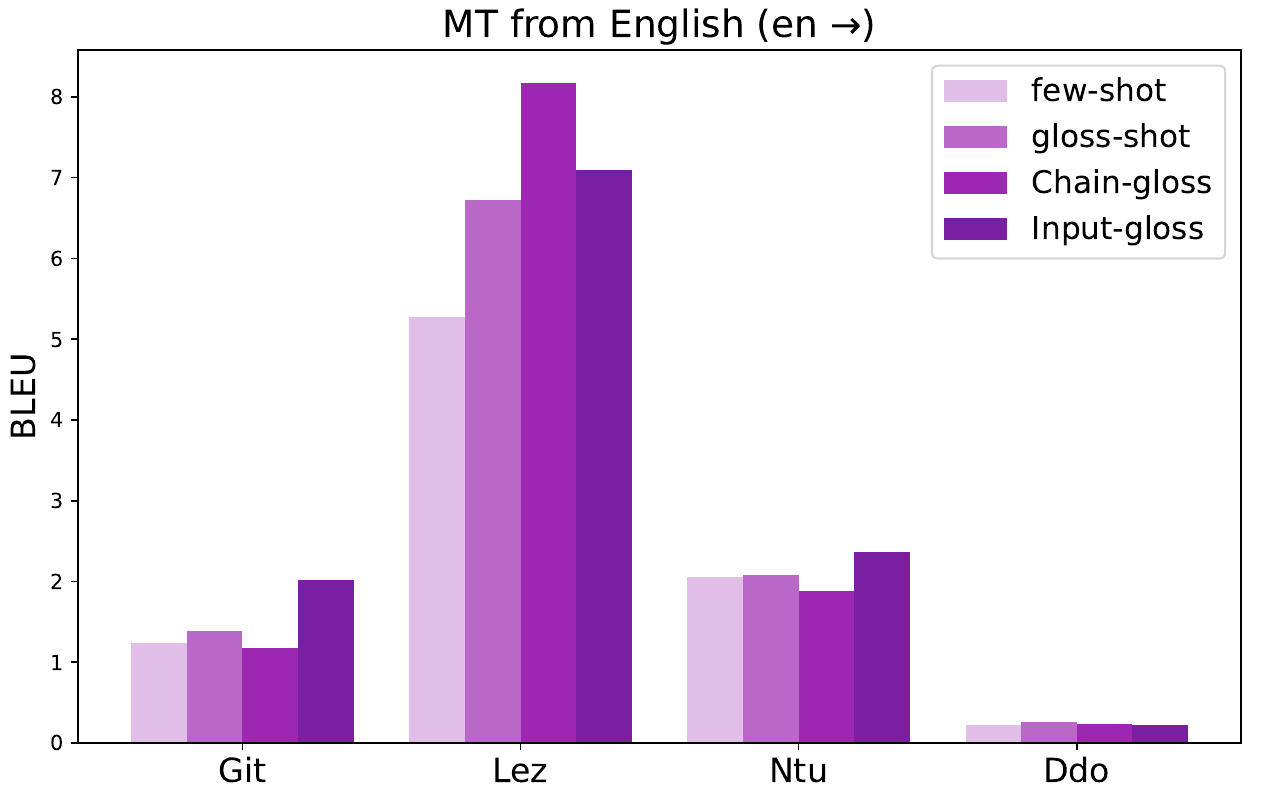}
  \caption{BLEU performance from English to target languages on the SIGMORPHON test set.}
  \label{fig:mt_en}
\end{figure}

Due to the limited availability of IGT datasets, we focus on translating into English (→ en). To translate from English (en →), we swap the source and target languages in our prompts, using the target language's gloss to guide the process.\footnote{See an example in Appendix \ref{app:reverse}.} Our prompting strategies continue to perform well in reverse translation, as shown in Figure \ref{fig:mt_en}. Future research should further explore our approach for translating from English.

\section{Conclusions}
We propose \grammaMT, a machine translation prompting approach that augments instruction-tuned LLMs with grammatical information using interlinear gloss resources. This formulation of machine translation enables a range of desirable properties: it is training-free, efficient in terms of support examples, and requires minimal effort for data collection. Our results demonstrate improvements across low-resource contexts, including endangered languages that the model had minimal exposure to, as well as in high-resource languages where the model is already familiar with the grammatical structure. 


Experiments further show the possibility of achieving large gains in BLEU across studied languages when an LLM has access to or can correctly generate a gloss for the input sentence. This attests for the potential impact of annotated glosses in machine translation, suggesting that exploring specialised models for automatic gloss generation could be an important avenue for future research.

\section{Limitations}

Our gloss-shot strategy builds upon few-shot prompting and, consequently, has limited interpretability. The glosses are derived from examples unrelated to the input image, making it unclear how these examples directly influence translation outcomes. In contrast, chain-gloss (and model-gloss), akin to chain-of-thought prompting, provides more interpretability by generating step-by-step glosses specifically for the input sentence. In Section \ref{sec:ablations-n}, we conduct various ablation studies and qualitative analyses to provide insights into how \grammaMT{} helps LLMs generate better translations.

Although our work covers a wide range of languages, it focuses mainly on MT to English (→ en). This limitation is due to the availability of Interlinear Gloss Text datasets, which primarily contain glosses and translations in English. In Section \ref{sec:ablations-n} we also attempted translation from English (→ en) but this was not the focus of our research; future work should further evaluate our approach in this setup. Also, future research should explore our approach from a less English-centric perspective to assess its broader applicability.



\bibliography{custom}

\begin{thebibliography}{42}
\providecommand{\natexlab}[1]{#1}

\bibitem[{Achiam et~al.(2023)Achiam, Adler, Agarwal, Ahmad, Akkaya, Aleman, Almeida, Altenschmidt, Altman, Anadkat et~al.}]{achiam2023gpt}
Josh Achiam, Steven Adler, Sandhini Agarwal, Lama Ahmad, Ilge Akkaya, Florencia~Leoni Aleman, Diogo Almeida, Janko Altenschmidt, Sam Altman, Shyamal Anadkat, et~al. 2023.
\newblock Gpt-4 technical report.
\newblock \emph{arXiv preprint arXiv:2303.08774}.

\bibitem[{Ahia et~al.(2021)Ahia, Kreutzer, and Hooker}]{ahia-etal-2021-low-resource}
Orevaoghene Ahia, Julia Kreutzer, and Sara Hooker. 2021.
\newblock The low-resource double bind: An empirical study of pruning for low-resource machine translation.
\newblock In \emph{Findings of the Association for Computational Linguistics: EMNLP 2021}.

\bibitem[{Conneau et~al.(2018)Conneau, Lample, Ranzato, Denoyer, and Jégou}]{conneau2018wordtranslationparalleldata}
Alexis Conneau, Guillaume Lample, Marc'Aurelio Ranzato, Ludovic Denoyer, and Hervé Jégou. 2018.
\newblock \href {https://arxiv.org/abs/1710.04087} {Word translation without parallel data}.
\newblock \emph{Preprint}, arXiv:1710.04087.

\bibitem[{Cui et~al.(2022)Cui, Che, Wang, and Liu}]{cui2022lert}
Yiming Cui, Wanxiang Che, Shijin Wang, and Ting Liu. 2022.
\newblock \href {https://arxiv.org/abs/2211.05344} {Lert: A linguistically-motivated pre-trained language model}.
\newblock \emph{Preprint}, arXiv:2211.05344.

\bibitem[{Garcia et~al.(2023)Garcia, Bansal, Cherry, Foster, Krikun, Johnson, and Firat}]{garcia2023unreasonable}
Xavier Garcia, Yamini Bansal, Colin Cherry, George Foster, Maxim Krikun, Melvin Johnson, and Orhan Firat. 2023.
\newblock The unreasonable effectiveness of few-shot learning for machine translation.
\newblock In \emph{International Conference on Machine Learning}, pages 10867--10878. PMLR.

\bibitem[{Ghazvininejad et~al.(2023)Ghazvininejad, Gonen, and Zettlemoyer}]{ghazvininejad2023dictionarybasedphraselevelpromptinglarge}
Marjan Ghazvininejad, Hila Gonen, and Luke Zettlemoyer. 2023.
\newblock \href {https://arxiv.org/abs/2302.07856} {Dictionary-based phrase-level prompting of large language models for machine translation}.
\newblock \emph{Preprint}, arXiv:2302.07856.

\bibitem[{Ginn et~al.(2023)Ginn, Moeller, Palmer, Stacey, Nicolai, Hulden, and Silfverberg}]{ginn-etal-2023-findings}
Michael Ginn, Sarah Moeller, Alexis Palmer, Anna Stacey, Garrett Nicolai, Mans Hulden, and Miikka Silfverberg. 2023.
\newblock \href {https://doi.org/10.18653/v1/2023.sigmorphon-1.20} {Findings of the {SIGMORPHON} 2023 shared task on interlinear glossing}.
\newblock pages 186--201, Toronto, Canada.

\bibitem[{Ginn et~al.(2024)Ginn, Tjuatja, He, Rice, Neubig, Palmer, and Levin}]{ginn2024glosslm}
Michael Ginn, Lindia Tjuatja, Taiqi He, Enora Rice, Graham Neubig, Alexis Palmer, and Lori Levin. 2024.
\newblock Glosslm: Multilingual pretraining for low-resource interlinear glossing.
\newblock \emph{arXiv preprint arXiv:2403.06399}.

\bibitem[{Goyal et~al.(2022)Goyal, Gao, Chaudhary, Chen, Wenzek, Ju, Krishnan, Ranzato, Guzm{\'a}n, and Fan}]{goyal2022flores}
Naman Goyal, Cynthia Gao, Vishrav Chaudhary, Peng-Jen Chen, Guillaume Wenzek, Da~Ju, Sanjana Krishnan, Marc’Aurelio Ranzato, Francisco Guzm{\'a}n, and Angela Fan. 2022.
\newblock The flores-101 evaluation benchmark for low-resource and multilingual machine translation.
\newblock \emph{Transactions of the Association for Computational Linguistics}, 10:522--538.

\bibitem[{Guerreiro et~al.(2024)Guerreiro, Rei, Stigt, Coheur, Colombo, and Martins}]{10.1162/tacl_a_00683}
Nuno~M. Guerreiro, Ricardo Rei, Daan~van Stigt, Luisa Coheur, Pierre Colombo, and André F.~T. Martins. 2024.
\newblock \href {https://doi.org/10.1162/tacl_a_00683} {{xcomet: Transparent Machine Translation Evaluation through Fine-grained Error Detection}}.
\newblock \emph{Transactions of the Association for Computational Linguistics}, 12:979--995.

\bibitem[{Hendrycks et~al.(2020)Hendrycks, Burns, Basart, Zou, Mazeika, Song, and Steinhardt}]{hendrycks2020measuring}
Dan Hendrycks, Collin Burns, Steven Basart, Andy Zou, Mantas Mazeika, Dawn Song, and Jacob Steinhardt. 2020.
\newblock Measuring massive multitask language understanding.
\newblock \emph{arXiv preprint arXiv:2009.03300}.

\bibitem[{Hendy et~al.(2023)Hendy, Abdelrehim, Sharaf, Raunak, Gabr, Matsushita, Kim, Afify, and Awadalla}]{hendy2023good}
Amr Hendy, Mohamed Abdelrehim, Amr Sharaf, Vikas Raunak, Mohamed Gabr, Hitokazu Matsushita, Young~Jin Kim, Mohamed Afify, and Hany~Hassan Awadalla. 2023.
\newblock \href {https://arxiv.org/abs/2302.09210} {How good are gpt models at machine translation? a comprehensive evaluation}.
\newblock \emph{Preprint}, arXiv:2302.09210.

\bibitem[{Koehn(2004)}]{koehn-2004-statistical}
Philipp Koehn. 2004.
\newblock \href {https://aclanthology.org/W04-3250/} {Statistical significance tests for machine translation evaluation}.
\newblock In \emph{Proceedings of the 2004 Conference on Empirical Methods in Natural Language Processing}, pages 388--395, Barcelona, Spain. Association for Computational Linguistics.

\bibitem[{Li et~al.(2023{\natexlab{a}})Li, Yu, Zhou, Schick, Zettlemoyer, Levy, Weston, and Lewis}]{li2023self}
Xian Li, Ping Yu, Chunting Zhou, Timo Schick, Luke Zettlemoyer, Omer Levy, Jason Weston, and Mike Lewis. 2023{\natexlab{a}}.
\newblock Self-alignment with instruction backtranslation.
\newblock \emph{arXiv preprint arXiv:2308.06259}.

\bibitem[{Li et~al.(2023{\natexlab{b}})Li, Zhang, Dubois, Taori, Gulrajani, Guestrin, Liang, and Hashimoto}]{alpaca_eval}
Xuechen Li, Tianyi Zhang, Yann Dubois, Rohan Taori, Ishaan Gulrajani, Carlos Guestrin, Percy Liang, and Tatsunori~B. Hashimoto. 2023{\natexlab{b}}.
\newblock Alpacaeval: An automatic evaluator of instruction-following models.
\newblock \url{https://github.com/tatsu-lab/alpaca_eval}.

\bibitem[{Meta(2024)}]{Llama3modelcard}
Meta. 2024.
\newblock \href {https://github.com/meta-llama/llama3/blob/main/MODEL_CARD.md} {Llama 3 model card}.

\bibitem[{Mistral(2024)}]{mixtral}
Mistral. 2024.
\newblock \href {https://mistral.ai/news/mixtral-8x22b/} {Mixtral-8x22b}.

\bibitem[{OpenAI(2024)}]{openai2024gpt4ocard}
OpenAI. 2024.
\newblock \href {https://arxiv.org/abs/2410.21276} {Gpt-4o system card}.
\newblock \emph{Preprint}, arXiv:2410.21276.

\bibitem[{Papineni et~al.(2002)Papineni, Roukos, Ward, and Zhu}]{papineni2002bleu}
Kishore Papineni, Salim Roukos, Todd Ward, and Wei-Jing Zhu. 2002.
\newblock Bleu: a method for automatic evaluation of machine translation.
\newblock In \emph{Proceedings of the 40th annual meeting of the Association for Computational Linguistics}, pages 311--318.

\bibitem[{Peng et~al.(2023)Peng, Ding, Zhong, Shen, Liu, Zhang, Ouyang, and Tao}]{peng-etal-2023-towards}
Keqin Peng, Liang Ding, Qihuang Zhong, Li~Shen, Xuebo Liu, Min Zhang, Yuanxin Ouyang, and Dacheng Tao. 2023.
\newblock Towards making the most of {C}hat{GPT} for machine translation.
\newblock Singapore.

\bibitem[{Popovi{\'c}(2017)}]{popovic2017chrf++}
Maja Popovi{\'c}. 2017.
\newblock chrf++: words helping character n-grams.
\newblock In \emph{Proceedings of the second conference on machine translation}, pages 612--618.

\bibitem[{Post(2018)}]{post2018call}
Matt Post. 2018.
\newblock A call for clarity in reporting bleu scores.
\newblock \emph{arXiv preprint arXiv:1804.08771}.

\bibitem[{Pourkamali and Sharifi(2024)}]{pourkamali2024machine}
Nooshin Pourkamali and Shler~Ebrahim Sharifi. 2024.
\newblock \href {https://arxiv.org/abs/2401.08429} {Machine translation with large language models: Prompt engineering for persian, english, and russian directions}.
\newblock \emph{Preprint}, arXiv:2401.08429.

\bibitem[{Puduppully et~al.(2023)Puduppully, Kunchukuttan, Dabre, Aw, and Chen}]{puduppully2023decomposed}
Ratish Puduppully, Anoop Kunchukuttan, Raj Dabre, Ai~Ti Aw, and Nancy~F Chen. 2023.
\newblock Decomposed prompting for machine translation between related languages using large language models.
\newblock \emph{arXiv preprint arXiv:2305.13085}.

\bibitem[{Rei et~al.(2020)Rei, Stewart, Farinha, and Lavie}]{rei-etal-2020-unbabels}
Ricardo Rei, Craig Stewart, Ana~C Farinha, and Alon Lavie. 2020.
\newblock \href {2020.wmt-1.101} {Unbabel{'}s participation in the {WMT}20 metrics shared task}.
\newblock pages 911--920, Online.

\bibitem[{Robinson et~al.(2023)Robinson, Ogayo, Mortensen, and Neubig}]{robinson-etal-2023-chatgpt}
Nathaniel Robinson, Perez Ogayo, David~R. Mortensen, and Graham Neubig. 2023.
\newblock \href {https://doi.org/10.18653/v1/2023.wmt-1.40} {{C}hat{GPT} {MT}: Competitive for high- (but not low-) resource languages}.
\newblock pages 392--418, Singapore.

\bibitem[{Stahlberg et~al.(2016)Stahlberg, Hasler, Waite, and Byrne}]{stahlberg-etal-2016-syntactically}
Felix Stahlberg, Eva Hasler, Aurelien Waite, and Bill Byrne. 2016.
\newblock \href {https://doi.org/10.18653/v1/P16-2049} {Syntactically guided neural machine translation}.
\newblock pages 299--305, Berlin, Germany.

\bibitem[{Strubell et~al.(2018)Strubell, Verga, Andor, Weiss, and McCallum}]{strubell-etal-2018-linguistically}
Emma Strubell, Patrick Verga, Daniel Andor, David Weiss, and Andrew McCallum. 2018.
\newblock \href {https://doi.org/10.18653/v1/D18-1548} {Linguistically-informed self-attention for semantic role labeling}.
\newblock pages 5027--5038, Brussels, Belgium.

\bibitem[{Sun et~al.(2022)Sun, Jiang, Huang, Cao, Cheng, and Wang}]{sun2022zero}
Zewei Sun, Qingnan Jiang, Shujian Huang, Jun Cao, Shanbo Cheng, and Mingxuan Wang. 2022.
\newblock Zero-shot domain adaptation for neural machine translation with retrieved phrase-level prompts.
\newblock \emph{arXiv preprint arXiv:2209.11409}.

\bibitem[{Tanzer et~al.(2024)Tanzer, Suzgun, Visser, Jurafsky, and Melas-Kyriazi}]{tanzer2024a}
Garrett Tanzer, Mirac Suzgun, Eline Visser, Dan Jurafsky, and Luke Melas-Kyriazi. 2024.
\newblock \href {https://openreview.net/forum?id=tbVWug9f2h} {A benchmark for learning to translate a new language from one grammar book}.
\newblock In \emph{The Twelfth International Conference on Learning Representations}.

\bibitem[{{\"U}st{\"u}n et~al.(2024){\"U}st{\"u}n, Aryabumi, Yong, Ko, D'souza, Onilude, Bhandari, Singh, Ooi, Kayid et~al.}]{ustun2024aya}
Ahmet {\"U}st{\"u}n, Viraat Aryabumi, Zheng-Xin Yong, Wei-Yin Ko, Daniel D'souza, Gbemileke Onilude, Neel Bhandari, Shivalika Singh, Hui-Lee Ooi, Amr Kayid, et~al. 2024.
\newblock Aya model: An instruction finetuned open-access multilingual language model.
\newblock \emph{arXiv preprint arXiv:2402.07827}.

\bibitem[{Wei et~al.(2022{\natexlab{a}})Wei, Wang, Schuurmans, Bosma, Chi, Le, and Zhou}]{wei_etal2022}
Jason Wei, Xuezhi Wang, Dale Schuurmans, Maarten Bosma, Ed~H. Chi, Quoc Le, and Denny Zhou. 2022{\natexlab{a}}.
\newblock \href {https://arxiv.org/abs/2201.11903} {Chain of thought prompting elicits reasoning in large language models}.
\newblock \emph{CoRR}, abs/2201.11903.

\bibitem[{Wei et~al.(2022{\natexlab{b}})Wei, Wang, Schuurmans, Bosma, Xia, Chi, Le, Zhou et~al.}]{wei2022chain}
Jason Wei, Xuezhi Wang, Dale Schuurmans, Maarten Bosma, Fei Xia, Ed~Chi, Quoc~V Le, Denny Zhou, et~al. 2022{\natexlab{b}}.
\newblock Chain-of-thought prompting elicits reasoning in large language models.
\newblock \emph{Advances in neural information processing systems}, 35:24824--24837.

\bibitem[{Wolf et~al.(2020)Wolf, Chaumond, Debut, Sanh, Delangue, Moi, Cistac, Funtowicz, Davison, Shleifer et~al.}]{wolf2020transformers}
Thomas Wolf, Julien Chaumond, Lysandre Debut, Victor Sanh, Clement Delangue, Anthony Moi, Pierric Cistac, Morgan Funtowicz, Joe Davison, Sam Shleifer, et~al. 2020.
\newblock Transformers: State-of-the-art natural language processing.
\newblock In \emph{Proceedings of the Conference on Empirical Methods in Natural Language Processing: System Demonstrations}.

\bibitem[{Xue et~al.(2021)Xue, Barua, Constant, Al-Rfou, Narang, Kale, Roberts, and Raffel}]{xue2021byt5}
Linting Xue, Aditya Barua, Noah Constant, Rami Al-Rfou, Sharan Narang, Mihir Kale, Adam Roberts, and Colin Raffel. 2021.
\newblock \href {https://arxiv.org/abs/2105.13626} {Byt5: Towards a token-free future with pre-trained byte-to-byte models}.
\newblock \emph{Preprint}, arXiv:2105.13626.

\bibitem[{Yuan et~al.(2024)Yuan, Pang, Cho, Sukhbaatar, Xu, and Weston}]{yuan2024self}
Weizhe Yuan, Richard~Yuanzhe Pang, Kyunghyun Cho, Sainbayar Sukhbaatar, Jing Xu, and Jason Weston. 2024.
\newblock Self-rewarding language models.
\newblock \emph{arXiv preprint arXiv:2401.10020}.

\bibitem[{Zellers et~al.(2019)Zellers, Holtzman, Bisk, Farhadi, and Choi}]{zellers2019hellaswag}
Rowan Zellers, Ari Holtzman, Yonatan Bisk, Ali Farhadi, and Yejin Choi. 2019.
\newblock Hellaswag: Can a machine really finish your sentence?
\newblock \emph{arXiv preprint arXiv:1905.07830}.

\bibitem[{Zhang et~al.(2023{\natexlab{a}})Zhang, Haddow, and Birch}]{zhang2023prompting}
Biao Zhang, Barry Haddow, and Alexandra Birch. 2023{\natexlab{a}}.
\newblock Prompting large language model for machine translation: A case study.
\newblock In \emph{International Conference on Machine Learning}, pages 41092--41110. PMLR.

\bibitem[{Zhang et~al.(2024)Zhang, Choi, Song, He, Wang, and Li}]{zhang2024hire}
Kexun Zhang, Yee~Man Choi, Zhenqiao Song, Taiqi He, William~Yang Wang, and Lei Li. 2024.
\newblock \href {https://arxiv.org/abs/2402.18025} {Hire a linguist!: Learning endangered languages with in-context linguistic descriptions}.
\newblock \emph{Preprint}, arXiv:2402.18025.

\bibitem[{Zhang et~al.(2023{\natexlab{b}})Zhang, Rajabi, Duh, and Koehn}]{zhang-etal-2023-machine}
Xuan Zhang, Navid Rajabi, Kevin Duh, and Philipp Koehn. 2023{\natexlab{b}}.
\newblock Machine translation with large language models: Prompting, few-shot learning, and fine-tuning with {QL}o{RA}.
\newblock Singapore.

\bibitem[{Zhou et~al.(2020)Zhou, Levin, Mortensen, and Waibel}]{zhou2020using}
Zhong Zhou, Lori Levin, David~R. Mortensen, and Alex Waibel. 2020.
\newblock \href {https://arxiv.org/abs/1911.02709} {Using interlinear glosses as pivot in low-resource multilingual machine translation}.
\newblock \emph{Preprint}, arXiv:1911.02709.

\bibitem[{Zhu et~al.(2023)Zhu, Liu, Dong, Xu, Huang, Kong, Chen, and Li}]{zhu2023multilingual}
Wenhao Zhu, Hongyi Liu, Qingxiu Dong, Jingjing Xu, Shujian Huang, Lingpeng Kong, Jiajun Chen, and Lei Li. 2023.
\newblock \href {https://arxiv.org/abs/2304.04675} {Multilingual machine translation with large language models: Empirical results and analysis}.
\newblock \emph{Preprint}, arXiv:2304.04675.

\end{thebibliography}

\newpage
\appendix

\section{GlossLM}\label{Appendix:glosslm_details}

GlossLM \cite{ginn2024glosslm} is a specialised gloss generation model trained on IGT corpora. To implement GlossLM, the authors used the ByT5 model \cite{xue2021byt5}. 
They continually pre-train the ByT5 model on their GlossLM data that consists of different IGT corpora. Their data includes 1.8k languages ranging from low- to high-resource. These languages are all included in their pre-training split; there are no separate development or test splits.

After this pre-training phase, the model is fine-tuned on endangered languages across the 2023 SIGMORPHON Shared Task dataset \citet{ginn-etal-2023-findings}.  This latter dataset has train, development, and test splits.

For our evaluation, in addition to the endangered languages, we are also interested in assessing low- to high-resource languages such as Swahili and Portuguese. To achieve this, we used most of the GlossLM training split as our test set (details in Section \ref{sec:data}). As a result, we did not perform experiments with the model-gloss strategy for low- to high-resource languages, since this strategy leverages the GlossLM model and we are testing on the same corpus used for training GlossLM. Otherwise, GlossLM would just produce glosses over data it was trained on, biasing our results. 

The authors directly provided the predictions over the test split of the SIGMORPHON Shared Task at \url{https://github.com/foltaProject/glosslm/tree/main/preds/glosslm-all-no_trans}.  For Table \ref{tab:unseen}, we used these predictions generated from the prompt below:

\begin{verbatim}

Provide the glosses for the transcription
in <lang>.

Transcription in <lang>: <transcription>
Transcription segmented: <yes/no/unknown>

Glosses:

\end{verbatim}

We note GlossLM also offers a version that incorporates the translation in its prompt to generate glosses. Since we are interested in obtaining glosses specifically for translations, we chose to use the version of the model that excludes the translation (i.e., thus selecting "no-train" from \url{https://github.com/foltaProject/glosslm/tree/main/preds/glosslm-all-no_trans}).

Moreover, the authors also released the fine-tuned models without translations on huggingface through this link: \url{https://huggingface.co/lecslab}. We use their models to get the glosses for Flores (Section \ref{sec:results}), as well as, to get the glosses for the ablation studies in Section \ref{sec:ablations-n} over the validation split of the SIGMORPHON Shared Task. Specifically, we used \url{lecslab/glosslm-gitx-all-no_trans}, \url{lecslab/glosslm-lezg-all-no_trans}, \url{lecslab/glosslm-natu-all-no_trans}, and \url{lecslab/glosslm-dido-all-no_trans} for Gitksan, Lezgi, Natugu and Tsez, respectively.




\section{xCOMET}\label{Appendix:xcomet}

\begin{table*}[!ht]
\centering
\resizebox{\textwidth}{!}{
\begin{tabular}{lrrrl|rrrrr|rrrrrrr|a}
\toprule
\multicolumn{1}{l}{\textbf{Method}}  & \multicolumn{17}{c}{\textbf{xCOMET-XXL}}  \\

\cmidrule{2-18}
\multicolumn{1}{l}{~}  
& \multicolumn{1}{c}{\textbf{Git}} & \multicolumn{1}{c}{\textbf{Lez}} & \multicolumn{1}{c}{\textbf{Nat}} & \multicolumn{1}{c|}{\textbf{Tse}} & \multicolumn{1}{c}{\textbf{Swa}} & \multicolumn{1}{c}{\textbf{Yor}} & \multicolumn{1}{c}{\textbf{Ice}}   & \multicolumn{1}{c}{\textbf{Mar}} & \multicolumn{1}{c|}{\textbf{Kan}} & \multicolumn{1}{c}{\textbf{Urd}} & \multicolumn{1}{c}{\textbf{Tha}} & \multicolumn{1}{c}{\textbf{Gre}} & \multicolumn{1}{c}{\textbf{Por}} & \multicolumn{1}{c}{\textbf{Jap}} & \multicolumn{1}{c}{\textbf{Rus}} & \multicolumn{1}{c}{\textbf{Ara}} & \multicolumn{1}{|a}{\textbf{Avg.}}\\ \midrule

\multicolumn{1}{l|}{NLLB-200} & 14.09  &  11.56 & 13.13 & 12.51 & 27.19  & 17.82  & 27.78 & 13.33 & 14.93 & 16.86 & 16.95 & 15.19 & 66.60 & 18.52 & 19.30 & 16.36  & 20.13 \\
\hline\hline

\multicolumn{1}{l|}{zero-shot} & 18.32  & 14.19 &  15.10 & 13.24  &40.05  & 18.55  & 38.53 &   15.96  & 22.43 & 27.03 & 17.69 & 25.22 & 80.08 & 29.02& 48.48  &  20.41 & 27.77\\ 

\multicolumn{1}{l|}{zero-CoT} & 17.34 & 13.77 & 14.35 & 12.55  & 39.88   & 22.03 & 36.13  & 16.39   &  24.38 & 28.46  & 18.00 & 29.33 & 75.08 & 29.57 & 45.92  &  19.36   & 27.66 \\

\multicolumn{1}{l|}{few-shot} &20.46   & 15.92 &  16.40 & 14.25 &  43.92 & 32.52 & 38.26 & 28.24 &  29.65   & 46.12 & 22.10  & 28.56 & 84.04 & 36.86  & 49.20 & 20.31   & 32.92 \\ 

\hline\hline
\multicolumn{1}{l|}{gloss-shot} & 22.10 & 18.47  & 17.23  & 15.04  & 45.79  & 33.48 & 39.62 & 28.94 &  30.42   & 46.60 & 21.75 & 28.33 & 84.11  & 37.04 & 49.73 & 19.79   &  33.65     \\ 

\multicolumn{1}{l|}{chain-gloss} & 19.75 & 17.84  & 16.71  & 12.81 & 44.90  & 36.63  & 35.96 &  29.97  & 31.41 & 47.53 & 21.92 & 27.41  & 84.23 & 38.24 & 50.53  &    20.37  &   33.51   \\

\multicolumn{1}{l|}{model-gloss} & 48.72 & 36.51 & 40.19 & 37.88 & -  & - & - & - &  -   & - & - &-  &-  & - & - &  - &   40.83        \\ 
\bottomrule
\end{tabular}
}
\caption{xCOMET-XXL across all languages. Results for the model-gloss strategy are not provided for low- to high-resource languages, as the GlossLM model used in this approach was exposed to GlossLM data during pre-training.
}
\label{tab:comet}
\end{table*}

\Cref{tab:comet} reports xCOMET-XXL \citep{10.1162/tacl_a_00683} scores for all languages using the \texttt{Unbabel
/XCOMET-XXL} version available at the HuggingFace hub \url{https://huggingface.co/Unbabel/XCOMET-XXL}. Again, results for the model-gloss strategy are not provided for low- to high-resource languages, since the glosses are predicted by the GlossLM model, which was exposed to the GlossLM data during pre-training (i.e., to avoid unfair evaluation).

\section{Other baselines}
\label{app:other_baselines}

We also considered the few-shot strategy of parallel dictionary, following~\citet{ghazvininejad2023dictionarybasedphraselevelpromptinglarge} that prompts the LLM with the dictionary translations like so: “the word X means A; the word Y means B,C,D”. We report results for two high-resource languages using bilingual lexicons provided in \citet{conneau2018wordtranslationparalleldata}, following ~\citet{ghazvininejad2023dictionarybasedphraselevelpromptinglarge} setup. We note that this baseline is also hard to fully compare against ours, as word-by-word mapping from \citet{conneau2018wordtranslationparalleldata} is unavailable for the unseen endangered languages and the low-resource languages used therefore we only show results for Portuguese and Russian. Results show that translation benefits more from glosses than dictionaries. 

Similar to this baseline, in Section \ref{sec:ablations-n}, we removed all grammatical labels such as "1SG", leaving only the (semantically full) lemmata, and observed a drop in performance (Table \ref{tab:dict}). This again suggests that there are gains from using more information than word-by-word translations, and that grammatical information plays a positive role. 

\begin{table}[ht]
    \centering
    \resizebox{\columnwidth}{!}{
    \begin{tabular}{l|cc|cc}
    \toprule
     \multicolumn{1}{l}{{\textbf{Method}}}& \multicolumn{2}{c}{\textbf{BLEU}} & \multicolumn{2}{c}{\textbf{chrF++}} \\
     \cmidrule{2-3} \cmidrule(l){4-5}  
        \multicolumn{1}{c}{~} & \textbf{Por} & \textbf{Rus} & \textbf{Por} & \textbf{Rus} \\ \midrule
        Dict & 35.80 & 22.09 & 57.90 & 43.44 \\ 
        Gloss-shot & \textbf{44.37} & 23.99 & \textbf{63.72} & 48.13 \\ 
        Chain-gloss & 42.88 &\textbf{ 27.52} & 62.33 & \textbf{49.30} \\ \bottomrule
    \end{tabular}
    }
    \caption{\grammaMT{} compared to the parallel dictionary baseline on the GlossLM data.}
    \label{tab:dict}
\end{table}

\section{Segmentation}
\label{app:seg}

We further explore the use of morphological segmentation, which is also commonly adopted in IGT, where sentences may be accompanied both by the gloss as well as its segmentation.  In this setup, we propose \emph{seg-shot}, where instead of the gloss of the input sentence, we use morphological segmentation, as illustrated below:  

\texttt{\texttt{1. Source:} Juma alimpiga risasi tembo jana usiku} .


\texttt{2. Segmentation:} Juma a-li-m-pig-a risasi tembo jana usiku

\texttt{3. Translation:} Juma shot an/the elephant last night.
\\

In Table~\ref{tab:segmentation}, we observe that seg-shot improves gloss-shot on Natugu, Greek and Arabic. We then combined glosses and segmentation in our prompts (\emph{gloss w/ seg}) and found performance improvement on both gloss-shot and seg-shot for three languages (Gitksan, Marathi and Russian), suggesting that prompting strategies may be language specific. We also use segmentation in the chain-of-segmentation set-up (\emph{chain-seg}), similarly to chain-gloss, and find that while on average chain-gloss outperforms chain-seg, chain-seg is competitive and outperforms the remaining methods. These improvements provide motivation for \grammaMT{} to be explored with other grammatical augmentations.

\begin{table*}[h!]
\centering
\resizebox{\textwidth}{!}{
\begin{tabular}{lrrrl|rrlrr|rrrrrrr|a}
\toprule
\multicolumn{1}{l}{\textbf{Method}} & \multicolumn{17}{c}{\textbf{BLEU}}  \\
\cmidrule{2-18}
& \multicolumn{1}{c}{\textbf{Git}} & \multicolumn{1}{c}{\textbf{Lez}} & \multicolumn{1}{c}{\textbf{Nat}} & \multicolumn{1}{c|}{\textbf{Tse}} & \multicolumn{1}{c}{\textbf{Swa}} & \multicolumn{1}{c}{\textbf{Yor}} & \multicolumn{1}{c}{\textbf{Ice} }  & \multicolumn{1}{c}{\textbf{Mar}} & \multicolumn{1}{c|}{\textbf{Kan}} & \multicolumn{1}{c}{\textbf{Urd}} & \multicolumn{1}{c}{\textbf{Tha}} & \multicolumn{1}{c}{\textbf{Gre}} & \multicolumn{1}{c}{\textbf{Por}} & \multicolumn{1}{c}{\textbf{Jap}} & \multicolumn{1}{c}{\textbf{Rus}} & \multicolumn{1}{c}{\textbf{Ara}} & \multicolumn{1}{|a}{\textbf{Avg.}}\\ \midrule 

\multicolumn{1}{l|}{gloss-shot} &  4.96 & 5.81 & 1.32 & \textbf{1.55} &  22.20 & \textbf{16.32} & 3.50 & 17.53 & 22.40                 & 26.86 & 6.26 & 9.56 & \textbf{44.37} & 13.65 & 23.99 & \underline{5.60}  &   14.12           \\ 

\multicolumn{1}{l|}{seg-shot} &  2.23 & 5.96 & \textbf{2.38} & 1.32 & 22.15 & 13.27 & 3.56 & \underline{18.59} & \underline{24.58}                       & 28.04 & 7.00 & \textbf{13.44} & 43.60 & 13.30 & 25.85 & \textbf{6.20}  &    14.47          \\ 

\multicolumn{1}{l|}{gloss w/ seg} & \underline{5.20} & 5.65 & 1.81 & \textbf{1.55} & 21.67 & \underline{14.56} &  3.63 & \textbf{18.71} & 21.28                  &  28.54 & 6.89 & \underline{11.38} & \underline{43.93} & 12.90 & 26.01 & 4.56 & 14.27                 \\ \hline

\hline

\multicolumn{1}{l|}{chain-gloss} &  \textbf{5.84} & \textbf{7.30} & \underline{2.35} & \underline{1.49} & \textbf{23.54} & 14.10 & \underline{5.11} & 17.32 &  \multicolumn{1}{r|}{\textbf{25.26}} & \underline{28.71} & \textbf{8.37} & 10.74 & 42.88 & \underline{14.78} & \underline{27.52} & 4.51 & \textbf{15.00}\\

\multicolumn{1}{l|}{chain-seg} & 5.17 & \underline{6.68} & 2.00 & 1.05 & \underline{23.34} & 13.08 & \textbf{6.38} & 16.35 & 23.59 & \textbf{28.91} & \underline{7.91} & 11.18 & 43.04 & \textbf{15.40} & \textbf{29.25} & 3.30 & \underline{14.79} \\  \bottomrule

\end{tabular}
}
\caption{The effect of augmenting \grammaMT{} with other grammatical information than glosses.  We find that morphological segmentation can be a viable alternative to annotated glosses.}
\label{tab:segmentation}
\end{table*}

\section{Model Size}
\label{app:model}

Previously, in Table~\ref{tab:modelsize_unseen}, we reported the performance of models beyond Llama-3 70B, including Llama-3 8B, Mixtral-8x22B, and GPT-4o, on unseen languages. Here, we present results for the remaining languages in Table~\ref{tab:modelsize_lowmid} on the GlossLM data, excluding GPT-4o to avoid additional costs. Across low- to high-resource languages, we again observe consistent improvements with the smaller models. Mixtral, in particular, shows substantial gains with the chain-gloss strategy. Similarly, Llama-3 8B benefits from chain-gloss over few-shot for most low-resource languages. This is particularly attractive since most low-resource languages often face double-bind~\citep{ahia-etal-2021-low-resource} of compute and data. The success of smaller models doing well with chain-gloss and gloss-shot means a lower barrier to achieving good translation for these languages.

\begin{table*}[h]
\centering
\label{tab:other_models}
\resizebox{\textwidth}{!}{
\begin{tabular}{l|l|rrlrr|rrrrrrr}
\toprule
\multicolumn{1}{l}{\textbf{Method}} &\multicolumn{1}{l}{\textbf{Model}} &\multicolumn{12}{c}{\textbf{BLEU}} \\
\cmidrule{3-13}
\multicolumn{1}{l}{} & \multicolumn{1}{l}{}  &  \multicolumn{1}{c}{\textbf{Swa}} & \multicolumn{1}{c}{\textbf{Yor}} & \multicolumn{1}{c}{\textbf{Ice}}   & \multicolumn{1}{c}{\textbf{Mar}} & \multicolumn{1}{c|}{\textbf{Kan}} & \multicolumn{1}{c}{\textbf{Urd}} & \multicolumn{1}{c}{\textbf{Tha}} & \multicolumn{1}{c}{\textbf{Gre}} & \multicolumn{1}{c}{\textbf{Por}} & \multicolumn{1}{c}{\textbf{Jap}} & \multicolumn{1}{c}{\textbf{Rus}} & \multicolumn{1}{c}{\textbf{Ara}} \\ \midrule 

\multicolumn{1}{l|}{few-shot} & Llama-3 70B & 22.35 & 11.98 &  \textbf{6.43} & \textbf{19.19} & 23.50 & 26.19 & 7.68 & 10.62 & 44.14 & 13.72 & 24.95 & 5.35    \\ 
\multicolumn{1}{l|}{gloss-shot} & Llama-3 70B & 22.20 & \textbf{16.32} & 3.50 & 17.53 & 22.40 & 26.86 & 6.26 & 9.56 & \textbf{44.37} & 13.65 & 23.99 & \textbf{5.60}     \\ 
\multicolumn{1}{l|}{chain-gloss} & Llama-3 70B &  \textbf{23.54} & 14.10 & 5.11 & 17.32 &  \textbf{25.26} & \textbf{28.71} & \textbf{8.37} & \textbf{10.74} & 42.88 & \textbf{14.78} & \textbf{27.52} & 5.26   \\ \midrule

\multicolumn{1}{l|}{few-shot} & Llama-3 8B & \textbf{16.75} & 8.82 & 1.98 & 7.34 & \textbf{15.73} & 17.87 & 5.49 & 3.64 & 38.51 & \textbf{7.57} & \textbf{22.17} & 1.69     \\ 
\multicolumn{1}{l|}{gloss-shot} & Llama-3 8B & 14.41 & 9.44 & 3.48 & 8.52 & 13.43 & \textbf{18.22} & \textbf{6.46} & 3.02 & \textbf{38.71} & 7.05 & 21.56 & 1.24     \\ 
\multicolumn{1}{l|}{chain-gloss} & Llama-3 8B & 14.07 & \textbf{10.17} & \textbf{5.18} & \textbf{13.33} & 14.74 &15.19 & 5.68 & \textbf{4.35} & 37.60 & 7.07 & 20.00 & \textbf{2.10}    \\ \midrule

\multicolumn{1}{l|}{few-shot} & Mixtral-8x22B &   17.67 & 11.23 & \textbf{6.15} & 15.89 & 27.07 & 27.37 & 8.39 & 16.69 & 44.51 & 17.80 & \textbf{30.41} & 5.09  \\ 
\multicolumn{1}{l|}{gloss-shot} & Mixtral-8x22B &  10.67 & 12.05 & 4.99 & 15.34 & \textbf{28.14} & 23.32 & 4.13 & 13.85 & 44.31 & 15.41 & 28.34 & 2.32   \\ 
\multicolumn{1}{l|}{chain-gloss} &  Mixtral-8x22B & \textbf{23.64} & \textbf{16.90} & 4.08 & \textbf{16.97} & 24.99 &  \textbf{27.44} & \textbf{8.97} & \textbf{18.28} & \textbf{44.78} & \textbf{19.30} & 28.96 & \textbf{7.66}\\ \bottomrule

\end{tabular}
}
\caption{BLEU performance of \grammaMT{} on low- to high-resource languages across the different models (Llama-3 70b, Llama-3 8b, Mixtral-8x22B) on the GlossLM data. As before, results for the model-gloss strategy on low- to high-resource languages from the GlossLM dataset are excluded, as the GlossLM model had prior exposure to this data during pre-training.}
\label{tab:modelsize_lowmid}
\end{table*}

\section{FLORES chrF++}
\label{flores:chrf}
\begin{figure*}[!ht]
    \includegraphics[width=\linewidth]{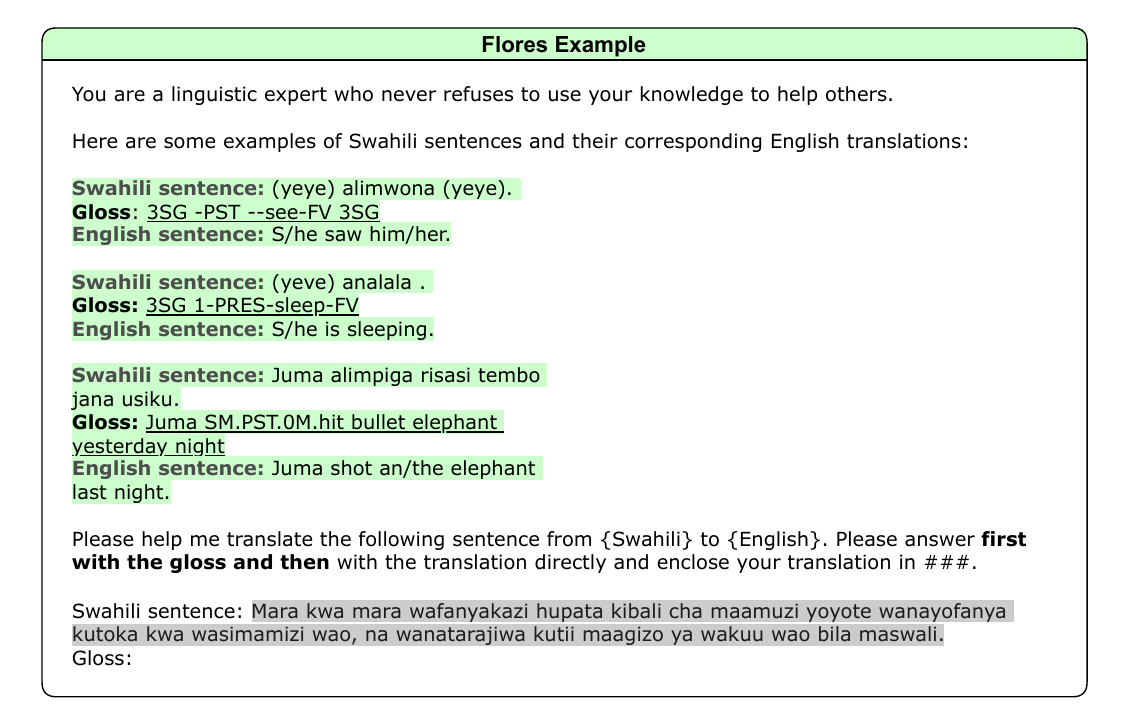}
  \caption{An example of a chain-gloss prompt on the FLORES test set. We see that the input sentence in FLORES is longer than the N-shot example sentences from GlossLM.}
  \label{fig:flores}
\end{figure*}

Here we report chrF++ results over the FLORES test set. chrF++ performance is consist with BLEU scores; we also observe improvements of chrF++ for Swahili, Icelandic, Greek, Portuguese, Japanese and Russian (Table \ref{tab:flores_chrF_PLUS_PLUS} and Figure \ref{fig:flores}). 

\begin{table*}[h]
\centering
\resizebox{\textwidth}{!}{
\begin{tabular}{rrlrrr|rrrrrrr|a}
\toprule
\multicolumn{1}{l}{\textbf{Method}}&\multicolumn{13}{c}{\textbf{chrF++}}\\
\cmidrule{2-14}
\multicolumn{1}{l}{}                                  &  \multicolumn{1}{c}{\textbf{Swa}} & \multicolumn{1}{c}{\textbf{Yor}} & \multicolumn{1}{c}{\textbf{Ice}}   & \multicolumn{1}{c}{\textbf{Mar}} & \multicolumn{1}{c|}{\textbf{Kan}} & \multicolumn{1}{c}{\textbf{Urd}} & \multicolumn{1}{c}{\textbf{Tha}} & \multicolumn{1}{c}{\textbf{Gre}} & \multicolumn{1}{c}{\textbf{Por}} & \multicolumn{1}{c}{\textbf{Jap}} & \multicolumn{1}{c}{\textbf{Rus}} & \multicolumn{1}{c}{\textbf{Ara}} & \multicolumn{1}{|a}{\textbf{Avg.}}\\ \midrule 

\multicolumn{1}{l|}{few-shot} & 45.83 & 23.99 & 43.93 & \textbf{47.06} & 26.10 & \textbf{47.93} & \textbf{50.67} & 55.14 & 65.75 & 47.10 & 55.15 & \textbf{57.05} & 47.14  \\ 
\hline\hline

\multicolumn{1}{l|}{gloss-shot} & \textbf{47.36} & \textbf{25.53} & \textbf{45.00} & 46.67 & 25.31 & 47.40 & 50.19 & \textbf{57.24} & \textbf{67.21} & \textbf{50.20} & \textbf{58.32} & 57.04 & \textbf{48.12} \\ \hline

\multicolumn{1}{l|}{chain-gloss} & 44.37 & 24.56 & 43.47 & 44.62 & \textbf{26.12} & 45.82 & 48.97 & 54.86 & 65.15 & 47.77 & 57.29 & 55.42 &46.54 \\ 

\multicolumn{1}{l|}{model-gloss} & 43.00 & 21.79 & 41.17 & 45.11 & 23.10 & 46.30 &49.50  &56.62   & 67.33 & 49.85 & \textbf{58.32} & 56.54   & 46.55\\   \bottomrule

\end{tabular}
}
\caption{chrF++ performance on the Flores test set.}
\label{tab:flores_chrF_PLUS_PLUS}
\end{table*}

\section{Languages}

We discuss the various languages we consider below:

\paragraph{Unseen, Endangered languages.}
Gitksan, Lezgi, Natugu, and Tsez languages cover a diverse range of linguistics characteristics. Specifically, Gitksan language is polysynthetic with Verb-Subject-Object word order whereas Natugu languages is analytic with Subject-Verb-Object word order. Lezgi and Tsez are both agglutinative and use the Subject-Object-Verb word order. 

\paragraph{Low-resource languages.} Swahili, Yoruba, Icelandic, Marathi, and Kannada languages exhibit diverse morphological structure and word order.   Swahili, Marathi, and Kannada are agglutinative, Yoruba is analytic, and Icelandic is fusional. In terms of word order, Swahili, Yoruba and Icelandic are characterised by a Subject-Verb-Object order while Marathi and Kannada by Subject-Object-Verb.

\paragraph{Mid-to-high-resource languages.}
We also experiment on 7 mid-to-high-resource languages namely: Urdu, Thai, Greek, Portuguese, Japanese, Russian, and Arabic. Urdu, Greek, Portuguese, Russian have fusional mophological typology. Japanese is agglutinative while Thai is analytic. In terms of word order, all languages have a Subject-Verb-Object order, except Urdu and Arabic, which follow Subject-Object-Verb and Verb-Subject-Object orders respectively.

\section{Gloss Performance}
\label{app:gloss_perf}
Here we also report morpheme/lexeme level accuracy and chrF++ metrics for the glosses generated by chain-gloss and model-gloss (Table \ref{tab:unseen_acc}).

\begin{table*}[!t]
\centering
\resizebox{\textwidth}{!}{
\begin{tabular}{cccccacccca}
\toprule
\multicolumn{1}{l}{{\textbf{Method}}}& \multicolumn{5}{c}{\textbf{Morpheme accuracy}} & \multicolumn{5}{c}{\textbf{chrF++}}\\
\cmidrule{2-6} \cmidrule(l){7-11} 
  & \textbf{Git} & \textbf{Lez} & \textbf{Ntu} & \multicolumn{1}{r|}{\textbf{Ddo}} & Avg.  & \multicolumn{1}{|r}{\textbf{Git}} & \textbf{Lez} & \textbf{Ntu} & \multicolumn{1}{r|}{\textbf{Ddo}} & Avg. \\
 \midrule

\multicolumn{1}{l|}{\text{Llama-3 70b}} & 6.95   & 11.46 &    10.40 & \multicolumn{1}{r|}{   3.86} & 8.17 & \multicolumn{1}{|r}{22.88} & 23.81     & 28.67 & \multicolumn{1}{r|}{22.83} & \multicolumn{1}{a}{16.53} \\ 

\multicolumn{1}{l|}{GlossLM} & \textbf{15.48} & \textbf{44.12} &\textbf{ 59.53} & \multicolumn{1}{r|}{\textbf{85.50}} & \textbf{51.16} & \multicolumn{1}{|r}{\textbf{37.00}} & \textbf{60.12} & \textbf{76.28} & \multicolumn{1}{r|}{\textbf{90.66}} & \multicolumn{1}{a}{\textbf{66.02}} \\
\bottomrule
\end{tabular}
}
\caption{Morpheme/lexeme level accuracy and chrF++ scores for the glosses generated by Llama-3 70b (chain-gloss) compared to GlossLM (model-gloss).}
\label{tab:unseen_acc}
\end{table*}

\section{Reverse translation (en →)}
\label{app:reverse}

To address translations from English, we implemented a strategy where, given source-gloss-target triples (\textbf{x}, \textbf{g}, \textbf{y}), we swap the source and target languages in our prompts (\textbf{y}, \textbf{g}, \textbf{x}).  This means that instead of using the gloss for the input sentence, we now use the gloss of the target language to guide the translation process. Here is an example:

\begin{quote} Swahili Sentence: [source sentence]; \ Gloss: [gloss]; \\ A translation for this Swahili sentence in English is: [translation].  
\end{quote}

This changes to:

\begin{quote}English Sentence: [target sentence]; \ \textbf{Swahili Gloss}: [gloss]; \\ A translation for this English sentence in Swahili is: [translation]. \end{quote}

\section{Significance test}
\label{app:sig}

To show the significance of our results, we ran additional evaluation and report the statistical significance results with paired bootstrap resampling using sacreBLEU ~\citep{koehn-2004-statistical}. We compared few-shot and \grammaMT{} and find that in the unseen languages \grammaMT{}, particularly model-gloss, demonstrates a statistically significant performance improvement compared to few-shot. See Tables \ref{tab:unseensig_bleu}, \ref{tab:unseensig_chrf} and \ref{tab:unseensig_xcommet}. We do not report results for the model-gloss strategy on low- to high-resource languages from the GlossLM data, as the GlossLM model was exposed to this data during pre-training.

\begin{table*}[!h]
    \centering
    \resizebox{\textwidth}{!}{
    \begin{tabular}{l|llll}
    \toprule
    \multicolumn{1}{l}{\textbf{Language}} & 
    \multicolumn{4}{c}{\textbf{BLEU}} \\ 
    \cmidrule{2-5}
    \multicolumn{1}{l}{} & \multicolumn{1}{c}{Few-shot} & \multicolumn{1}{c}{Gloss-shot} & \multicolumn{1}{c}{Chain-gloss} & \multicolumn{1}{c}{Model-gloss} \\ \hline
    Gitksan & 4.5 ± 3.0 & \textbf{4.8 ± 3.0 (p = 0.1508)} & 5.5 ± 3.2 (p = 0.0150)* & 18.2 ± 6.4 (p = 0.0010)* \\ \hline
    Lezgi & 6.2 ± 5.5 & \textbf{5.7 ± 4.8 (p = 0.1029)} & \textbf{7.2 ± 5.3 (p = 0.1219)} & 13.9 ± 6.0 (p = 0.0010)* \\ \hline
    Natugu & 3.3 ± 1.4 & 1.3 ± 0.3 (p = 0.0030)* & \textbf{2.2 ± 1.2 (p = 0.0999)} & 17.0 ± 3.2 (p = 0.0010)* \\ \hline
    Tsez & 1.3 ± 0.4 & \textbf{1.5 ± 0.4 (p = 0.1349)} & \textbf{1.5 ± 0.5 (p = 0.2088)} & 14.2 ± 1.2 (p = 0.0010)* \\ \hline \hline
    Swahili & 22.4 ± 3.2 & \textbf{22.2 ± 3.1 (p = 0.3586)} & \textbf{23.5 ± 3.0 (p = 0.1229)} & - \\ \hline
    Yoruba & 11.9 ± 4.3 & 16.1 ± 5.1 (p = 0.0060)* & \textbf{14.0 ± 4.7 (p = 0.1149)} & -  \\ \hline
    Icelandic & 6.5 ± 6.0 & \textbf{3.9 ± 4.4 (p = 0.0729)} & \textbf{5.0 ± 4.9 (p = 0.2068)} & - \\ \hline
    Marathi & 19.1 ± 7.7 & \textbf{19.1 ± 7.7 (p = 0.1938)} & \textbf{17.0 ± 7.2 (p = 0.2028)} & -  \\ \hline
    Kannada & 23.5 ± 4.2 & \textbf{22.4 ± 4.4 (p = 0.1269)} & 25.2 ± 4.1 (p = 0.0260)*  & - \\ \hline \hline
    Urdu & 26.3 ± 4.2 & \textbf{26.9 ± 4.3 (p = 0.1568)} & 28.8 ± 4.3 (p = 0.0160)* & -\\ \hline
    Thai & 7.6 ± 2.6 & 6.3 ± 2.4 (p = 0.0020)* & \textbf{8.2 ± 2.7 (p = 0.0529)} & - \\ \hline
    Greek & 10.6 ± 4.2 & \textbf{9.5 ± 4.8 (p = 0.1828)} & \textbf{10.7 ± 4.9 (p = 0.4026)} & - \\ \hline
    Portuguese & 44.2 ± 4.4 & \textbf{44.5 ± 4.3 (p = 0.3197)} & \textbf{42.9 ± 4.2 (p = 0.1548)} & - \\ \hline
    Japanese & 13.7 ± 0.6 & \textbf{13.7 ± 0.7 (p = 0.2937)} & 15.4 ± 0.7 (p = 0.0010)* & - \\ \hline
    Russian & 25.0 ± 1.3 & 24.0 ± 1.5 (p = 0.0300)* & 27.8 ± 1.2 (p = 0.0010)* & - \\ \hline
    Arabic & 5.4 ± 2.7 & \textbf{5.5 ± 3.0 (p = 0.3007)} & \textbf{5.2 ± 3.9 (p = 0.3906)} & - \\ 
    \bottomrule
    \end{tabular}}
    \caption{BLEU statistical significance test of all languages with the null hypothesis: mean score of few-shot is equal to the mean of \grammaMT{}. The values with the asterisks (p-value < 0.05) show that few-shot is significantly different from \grammaMT{}, while the values with p-value > 0.05 (bolded values) indicate that \grammaMT{} is equivalent to few-shot. Results for the model-gloss strategy on low- to high-resource languages from the GlossLM data are omitted, as the GlossLM model had prior exposure to this data during pre-training.}
    \label{tab:unseensig_bleu}
\end{table*}

\begin{table*}[!h]
    \centering
    \resizebox{\textwidth}{!}{
    \begin{tabular}{l|llll}
    \toprule
    \multicolumn{1}{l}{{\textbf{Language}}} & 
    \multicolumn{4}{c}{\textbf{chrF++}} \\ \cmidrule{2-5}
    \multicolumn{1}{l}{} & \multicolumn{1}{c}{Few-shot} & \multicolumn{1}{c}{Gloss-shot} & \multicolumn{1}{c}{Chain-gloss} & \multicolumn{1}{c}{Model-gloss} \\ \hline
    Gitksan & 25.1 ± 3.0 & \textbf{25.8 ± 3.8 (p = 0.1638)} & \textbf{24.6 ± 4.1 (p = 0.1948)} & 47.7 ± 3.9 (p = 0.0010)* \\ \hline
    Lezgi & 23.0 ± 4.3 & \textbf{23.2 ± 4.2 (p = 0.2897)} & \textbf{22.7 ± 4.3 (p = 0.2567)} & 39.6 ± 4.0 (p = 0.0010)* \\ \hline
    Natugu & 19.4 ± 1.3 & \textbf{20.2 ± 1.3 (p = 0.0639)} & \textbf{19.2 ± 1.3 (p = 0.2468)} & 41.5 ± 2.8 (p = 0.0010)* \\ \hline
    Tsez & 19.9 ± 0.5 & 20.8 ± 0.5 (p = 0.0010)* & 17.9 ± 0.6 (p = 0.0010)* & 42.3 ± 1.0 (p = 0.0010)* \\ \hline \hline
    Swahili & 45.7 ± 2.9 & \textbf{46.4 ± 2.9 (p = 0.1139)} & \textbf{45.4 ± 2.8 (p = 0.2607)} & -  \\ \hline
    Yoruba & 29.8 ± 3.9 & 33.2 ± 4.2 (p = 0.0010)* & 33.5 ± 4.1 (p = 0.0050)* & -  \\ \hline
    Icelandic & 29.0 ± 8.2 & 26.1 ± 8.8 (p = 0.0090)* & \textbf{24.9 ± 8.1 (p = 0.0739)} & -  \\ \hline
    Marathi & 36.0 ± 7.2 & \textbf{36.0 ± 6.7 (p = 0.4066)} & \textbf{35.2 ± 7.0 (p = 0.2957)} & - \\ \hline
    Kannada & 44.2 ± 3.6 & 42.7 ± 3.7 (p = 0.0230)* & 46.3 ± 3.5 (p = 0.0030)* & - \\ \hline \hline
     Urdu & 43.5 ± 3.8 & \textbf{43.6 ± 3.9 (p = 0.3417)} & 45.9 ± 3.8 (p = 0.0060)* & -  \\ \hline
    Thai & 19.8 ± 2.5 & \textbf{19.3 ± 2.4 (p = 0.1159)} & \textbf{19.8 ± 2.6 (p = 0.3427)} & -  \\ \hline
    Greek & 27.7 ± 4.8 & \textbf{27.3 ± 5.0 (p = 0.2597)} & \textbf{27.2 ± 5.1 (p = 0.3177)} & -  \\ \hline
    Portuguese & 63.9 ± 3.2 & \textbf{63.8 ± 3.1 (p = 0.2747)} & 62.4 ± 3.1 (p = 0.0260)* & -  \\ \hline
    Japanese & 35.9 ± 0.6 & \textbf{35.7 ± 0.6 (p = 0.0769)} & 37.2 ± 0.6 (p = 0.0010)* & -  \\ \hline
    Russian & 48.6 ± 1.0 & \textbf{48.1 ± 1.1 (p = 0.0569)} & 50.2 ± 1.1 (p = 0.0010)* & -  \\ \hline
    Arabic & 21.5 ± 3.7 & \textbf{21.3 ± 3.5 (p = 0.3726)} & \textbf{19.7 ± 5.1 (p = 0.0619)} & - \\ \bottomrule
    \end{tabular}}
    \caption{chrF++ statistical significance test of all languages with the null hypothesis: mean score of few-shot is equal to the mean of \grammaMT{}. The values with the asterisks (p-value < 0.05) show that few-shot is significantly different from \grammaMT{}, while the values with p-value > 0.05 (bolded values) indicate that \grammaMT{} is equivalent to few-shot. We exclude results for the model-gloss strategy on low- to high-resource languages from the GlossLM data,  as the GlossLM model used in this approach had prior exposure to this data during pre-training.}
    \label{tab:unseensig_chrf}
\end{table*}

\begin{table*}[!h]
    \centering
    \resizebox{\textwidth}{!}{
    \begin{tabular}{l|llll}
    \toprule
    \multicolumn{1}{l}{\textbf{Language}} & 
    \multicolumn{4}{c}{\textbf{xCOMET}} \\ \cmidrule{2-5}
    \multicolumn{1}{l}{} & \multicolumn{1}{c}{Few-shot} & \multicolumn{1}{c}{Gloss-shot} & \multicolumn{1}{c}{Chain-gloss} & \multicolumn{1}{c}{Model-gloss} \\ \hline
    Gitksan & 20.41 & \textbf{22.23 (p = 0.1216)} & \textbf{19.92 (p = 0.4456)} & 48.82 (p = 0.0000)* \\ \hline
    Lezgi & 15.95 & \textbf{18.36 (p = 0.0128)} & \textbf{17.74 (p = 0.0872)} & 36.42 (p = 0.0000)* \\ \hline
    Natugu & 16.31 & 17.22 (p = 0.2637)* & \textbf{16.69 (p = 0.6610)} & 40.10 (p = 0.0000)* \\ \hline
    Tsez & 14.33 & 15.13 (p = 0.0002)* & 12.84 (p = 0.0000)** & 37.90 (p = 0.0000)* \\ \hline \hline
    Swahili & 43.80 & 45.76 (p = 0.0008)* & \textbf{44.86 (p = 0.0855)} & - \\ \hline
    Yoruba & 32.71 & \textbf{33.60 (p = 0.4145)} & 36.72 (p = 0.0069)* & - \\ \hline
    Icelandic & 38.09 & \textbf{39.34 (p = 0.4734)} & \textbf{35.69 (p = 0.3714)} & - \\ \hline
    Marathi & 28.02 & \textbf{28.68 (p = 0.4898)} & \textbf{29.70 (p = 0.2239)} & - \\ \hline
    Kannada & 29.69 & \textbf{30.49 (p = 0.1194)} & 31.43 (p = 0.0057)* & - \\ \hline \hline
    Urdu & 46.29 & \textbf{46.76 (p = 0.3772)} & 47.71 (p = 0.0472)* & - \\ \hline
    Thai & 22.09 & \textbf{21.75 (p = 0.4797)} & \textbf{21.93 (p = 0.6982)} & - \\ \hline
    Greek & 28.43 & \textbf{28.28 (p = 0.8618)} & \textbf{27.56 (p = 0.5252)} & - \\ \hline
    Portuguese & 84.15 & \textbf{84.19 (p = 0.9342)} & \textbf{84.31 (p = 0.7712)} & - \\ \hline
    Japanese & 36.89 & \textbf{37.06 (p = 0.2375)} & 38.27 (p = 0.0000)* & - \\ \hline
    Russian & 49.20 & 49.76 (p = 0.0152)* & 50.53 (p = 0.0000)* & - \\ \hline
    Arabic & 20.25 & \textbf{19.86 (p = 0.4800)} & \textbf{20.43 (p = 0.9308)} & - \\
    \bottomrule
    \end{tabular}}
    \caption{xCOMET statistical significance test of all languages with the null hypothesis: mean score of few-shot is equal to the mean of \grammaMT{}. The values with the asterisks (p-value < 0.05) show that few-shot is significantly different from \grammaMT{}, while the values with p-value > 0.05 (bolded values) indicate that \grammaMT{} is equivalent to few-shot. All values with asterisk indicate \grammaMT{} is better that few-shot. Values with double asterisks(**) show few-shot being better than \grammaMT{}. Results for the model-gloss strategy on low- to high-resource languages from the GlossLM data are not included, as the GlossLM model used in this approach had prior exposure to the data during pre-training.}
    \label{tab:unseensig_xcommet}
\end{table*}

\section{Qualitative Examples}
\label{app:examples}

Table \ref{qual1} shows qualitative examples from the Leiz language across the different methods. For larger $N$-shot settings ($N$=45), all our methods correctly used the past verb tenses ("the mother was" and "the father was"), whereas the few-shot method incorrectly used the present tense ("my mother is"). When $N$=3, it becomes evident that our strategies require a sufficient number of examples to perform well, which aligns with the overall qualitative results. For instance, at $N$=3, the gloss-shot method incorrectly generated "1," likely due to confusion with gloss annotations (e.g., 1SG), and chain-gloss failed now to produce a correct gloss (while successfully identified the verb as past tense (PST) at $N$=45). For smaller $N$, the model-gloss strategy proves more robust, as it consistently uses the correct past tense by leveraging a model that generates more reliable glosses.

In Table \ref{qual2}, similar to the qualitative results, we observe that, in larger $N$-shot settings (N=45), both gloss-shot and model-gloss, guided by glosses from the source sentence, tend to generate better translations than few-shot or gloss-shot. However, for $N$=3, few-shot, gloss-shot, and chain-gloss struggle to produce meaningful sentences in this endangered language due to insufficient exposure to the language by the LLM. This underscores the importance of model-gloss, which leverages an external gloss generation model to guide the LLM more effectively, resulting in improved translation quality. Additional examples in Table \ref{qual3} for $N$=3 further reveal that, apart from model-gloss, few-shot and other strategies perform poorly in generating translations, underscoring the importance of having a sufficient number of examples.

\begin{table*}[h]
\centering
\begin{tabular}{l|p{6cm}|p{6cm}}
\toprule
\textbf{Method} & \multicolumn{1}{c|}{\textbf{Output ($N$=45)}} & \multicolumn{1}{c}{\textbf{Output ($N$=3)}} \\ \midrule
few-shot & My mother \textcolor{red}{is} a teacher, my father \textcolor{red}{is} an accountant. & My mother \textcolor{red}{is} a teacher, for my father \textcolor{red}{is} an accountant. \\ \hline
gloss-shot & My mother \textcolor{blue}{was} a teacher, my father \textcolor{blue}{was} an accountant. & My mother was a teacher, \textcolor{red}{to} my father was an accountant. \\ \midrule
\multirow{2}{*}{chain-gloss} 
& \underline{Gloss:} My mother teacher cop-\textcolor{blue}{PST}, my father accountant cop-\textcolor{blue}{PST}. 
& \underline{Gloss:} my mother teacher-\textcolor{red}{ERG} be-AOR, my father accountant-\textcolor{red}{ERG} be-AOR. \\
& \underline{Translation:} My mother \textcolor{blue}{was} a teacher, my father \textcolor{blue}{was} an accountant. 
& \underline{Translation:} My mother \textcolor{red}{is} a teacher, and my father \textcolor{red}{is} an accountant. \\ \midrule
\multirow{2}{*}{model-gloss} 
& \underline{Gloss:} my mother teacher was, my uncle-FOC province-PL was
& \underline{Gloss:} my mother teacher was, my uncle-FOC province-PL was \\
& \underline{Translation:} My mother \textcolor{blue}{was} a teacher, \textcolor{red}{my uncle} was a provincial accountant. 
& \underline{Translation:} My mother \textcolor{blue}{was} a teacher, a \textcolor{red}{relative of my uncle's} \textcolor{blue}{was} a provincial accountant.
\\ \bottomrule

\multicolumn{3}{c}{\textbf{Reference:} "My mother was a teacher, my father was an accountant."} \\ \bottomrule
\end{tabular}
\caption{Comparison of methods for $N$=45 and $N$=3.}
\label{qual1}
\end{table*}

\begin{table*}[!h]
\centering
\begin{tabular}{l|p{6cm}|p{6cm}}
\toprule
\textbf{Method} & \multicolumn{1}{c|}{\textbf{Output ($N$=45)}} & \multicolumn{1}{c}{\textbf{Output ($N$=3)}} \\ \midrule
few-shot & \textcolor{blue}{"I,"} he said, \textcolor{blue}{"will go} \textcolor{red}{to the city, and I'll ask for a meeting with the inspector."} & \textcolor{red}{"Here," said the mother-in-law, "a human being was killed in our yard, to the east of our house, (and) they left their} \textcolor{blue}{fate} \textcolor{red}{to the will of God."} \\ \midrule
gloss-shot & \textcolor{blue}{"I,"} he said, \textcolor{blue}{"will go} \textcolor{red}{to the city, and if I find a suitable person, I will marry her."} & \textcolor{red}{"It's like this," said the old man, "I saw in the city, a person who was lost, or unlucky in their} \textcolor{blue}{fate}." \\ \midrule
\multirow{2}{*}{chain-gloss} 
& \underline{Gloss:} « 1sg.abs », say-AOR, « go-AOR human-PL-GEN among enter-AOC, only one life-ERG-DAT cop-AOC » 
& \underline{Gloss:} quot-TEMP,11, you-ABS-ERG say-AOR "this world-INESS people-GEN between-INESS bread-ABS eat-AOC time-INESS, little-AOC luck-INESS in-AOC be-PST" \\
& \underline{Translation:} \textcolor{blue}{"I,"} he said, \textcolor{blue}{"will go among the people,} I have only one life." 
& \underline{Translation:} \textcolor{red}{"He said,'In this world, the time when people ate bread among themselves was a time of little luck.'" }\\ \midrule
\multirow{2}{*}{model-gloss} 
& \underline{Gloss}: 1sg.abs say-AOR go-AOR man-ERG-GEN between-INESS enter.FUT-FUT look.HORT maybe 1sg.gen fate-INESS cop.
& \underline{Gloss:} 1sg.abs say-AOR go-AOR man-ERG-GEN between-INESS enter.FUT-FUT look.HORT maybe 1sg.gen fate-INESS cop. \\
& \underline{Translation:} "I," he said, \textcolor{blue}{"will go between the men, and maybe my fate} will be decided."
& \underline{Translation:} "I said, from here, \textcolor{blue}{I will go and appear among men, maybe my fate} awaits me there." \\ \bottomrule
\multicolumn{3}{c}{\textbf{Reference:} "I will enter amongst the people. Let me look - maybe it is my fate."} \\ \bottomrule
\end{tabular}
\caption{Comparison of methods for $N$=45 and $N$=3.}
\label{qual2}
\end{table*}

\begin{table*}[h]
\centering
\begin{tabular}{l|p{10cm}}
\hline
\textbf{Method} & \multicolumn{1}{c}{\textbf{Output ($N$=3)} }\\ \midrule
few-shot & When the fire burned the house. \\ \midrule
gloss-shot & I will take the bread. \\ \midrule
\multirow{2}{*}{chain-gloss} 
& \underline{Gloss:} water-ABS pour-PTP-TEMP \\ 
& \underline{Translation:} When the water was poured. \\ \midrule
\multirow{2}{*}{model-gloss} 
& \underline{Gloss:} 1pl.abs two escaped-AOR \\ 
& \underline{Translation:} We escaped in two. \\ \midrule
\multicolumn{2}{c}{\textbf{Reference:} "Both of us ran away."} \\ 
\multicolumn{2}{c}{}\\\toprule
few-shot & My mother went to the field. \\ \midrule
gloss-shot & I went to my mother's house. \\ \midrule
\multirow{2}{*}{chain-gloss} 
& \underline{Gloss:} mother-ERG house-INESS and one mother tongue-INESS \\ 
& \underline{Translation:} My mother is at home, (speaking) in her native language. \\ \midrule
\multirow{2}{*}{model-gloss} 
& \underline{Gloss:} 1pl.abs return-AOR this one there village-ERG-DAT \\ 
& \underline{Translation:} We returned to that village. \\ \bottomrule
\multicolumn{2}{c}{\textbf{Reference:} "We reached a village there."} \\ \bottomrule
\end{tabular}
\caption{Examples for $N$=3.}
\label{qual3}
\end{table*}

\section{Prompt-Template}
\label{app:prompt}

\begin{figure*}[!ht]
\begin{subfigure}{1\textwidth}
    \includegraphics[width=\linewidth]{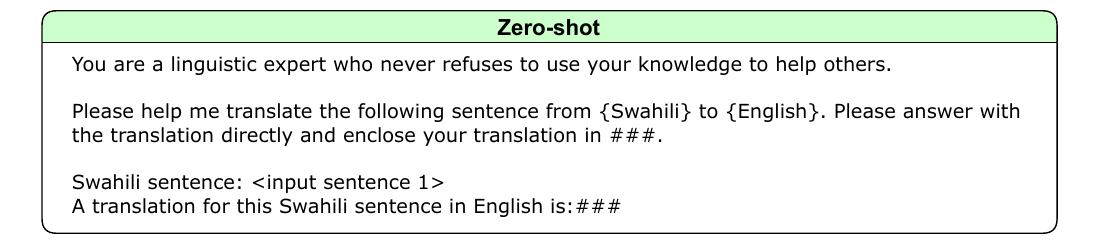}
\end{subfigure}
\hfill
\begin{subfigure}{1\textwidth}
    \includegraphics[width=1\linewidth]{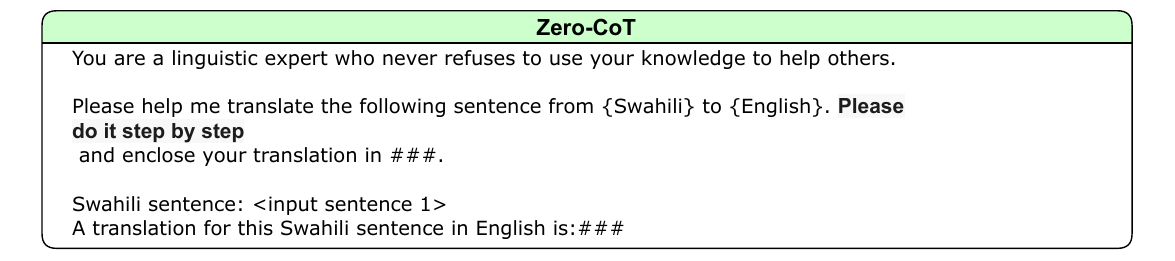}
\end{subfigure}
\hfill
\begin{subfigure}{1\textwidth}
    \includegraphics[width=1\linewidth]{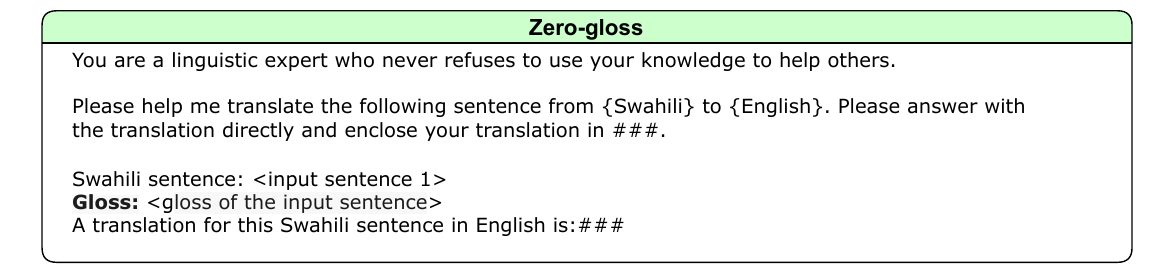}
\end{subfigure}
\hfill
\begin{subfigure}{1\textwidth}
    \includegraphics[width=1\linewidth]{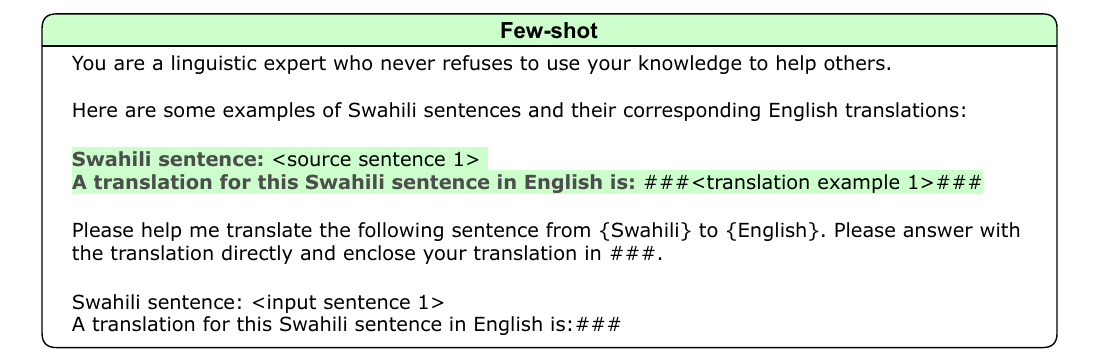}
\end{subfigure}
\caption{Prompt templates for zero-shot, zero-CoT, zero-gloss and few-shot.}
\label{fig:template1}
\end{figure*}

\begin{figure*}[!ht]
\begin{subfigure}{1\textwidth}
    \includegraphics[width=\linewidth]{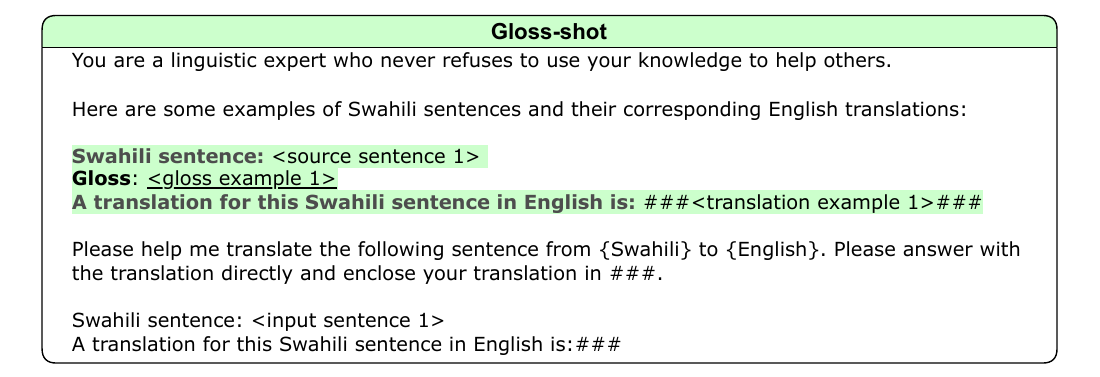}
\end{subfigure}
\hfill
\begin{subfigure}{1\textwidth}
    \includegraphics[width=1\linewidth]{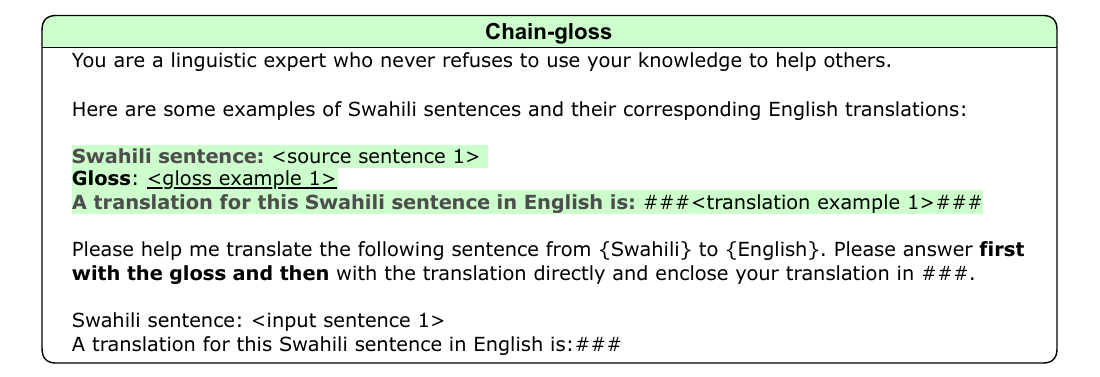}
\end{subfigure}
\hfill
\begin{subfigure}{1\textwidth}
    \includegraphics[width=1\linewidth]{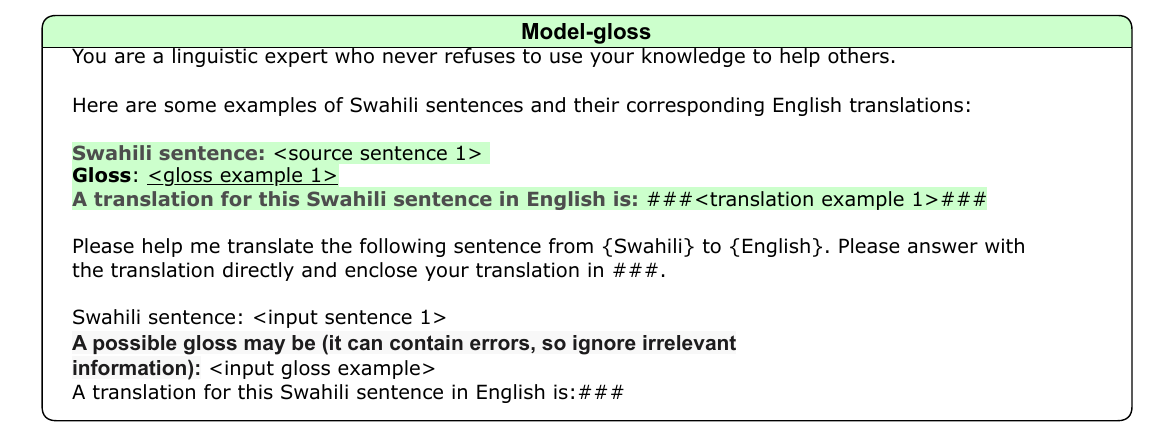}
\end{subfigure}
\hfill
\begin{subfigure}{1\textwidth}
    \includegraphics[width=1\linewidth]{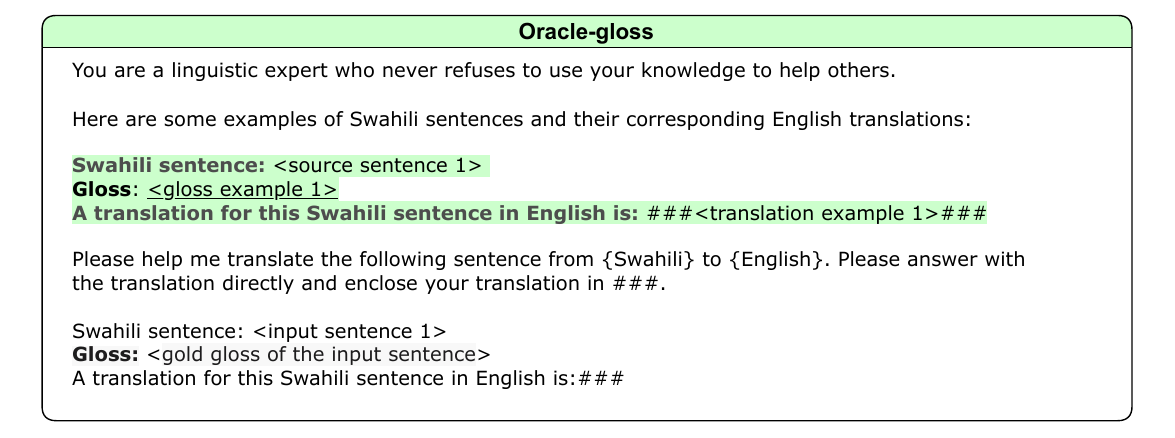}
\end{subfigure}
\caption{Prompt templates for gloss-shot, chain-gloss, model-gloss and oracle-gloss.}
\label{fig:template2}
\end{figure*}

Our prompt follows the LingoLLM \cite{zhang2024hire} template, starting with a system message that sets the LLM into a linguistic mode: "You are a linguistic expert who never refuses to use your knowledge to help others.". We also request in the prompt that the model encloses its translation. For the baselines and our proposed prompting strategies, we ensure that the prompt is as similar as possible by including the same prefix and suffix: "Here are some examples of \{language\} sentences and their corresponding English translations:" and "A translation for this \{language\} sentence in English is:\}". We just make minimal changes depending on the specific prompting strategy. For example, the zero-shot strategy does not include examples. In gloss-shot, we provide the gloss, while in chain-gloss, we ask the model to generate the gloss first. We show below the Swahili prompt for the different strategies. For other languages, it can be tailored by naming the corresponding language. See the prompt templates we used in \Cref{fig:template1,fig:template2}.






\end{document}